%% file: main.tex
\documentclass[letterpaper]{article} 
\usepackage[preprint]{aaai2027}
\usepackage[hyphens]{url}  
\usepackage{graphicx} 
\usepackage{natbib}  
\usepackage{caption} 
\usepackage{algorithm}
\usepackage{algorithmic}
\usepackage{newfloat}
\usepackage{amsfonts}
\usepackage{listings}
\usepackage{soul}
\usepackage{multirow}
\usepackage{pifont}

\usepackage{todonotes}

\usepackage{newfloat}
\usepackage{listings}
\DeclareCaptionStyle{ruled}{labelfont=normalfont,labelsep=colon,strut=off} 
\floatstyle{ruled}
\newfloat{listing}{tb}{lst}{}
\floatname{listing}{Listing}

\usepackage{booktabs}

\input{math_symbols}

\newcommand{\bh}{\mathbf{h}}

\title{Breaking Chains with Trees: Model-Parallel Deep Learning with $\mathcal{O}(\log N)$ Time Complexity
}

\author{
\small
Neeraj Mohan Sushma\textsuperscript{\rm 1,2},
Aditya Nagarsekar\textsuperscript{\rm 3},
Cabrel Teguemne Fokam\textsuperscript{\rm 2},
Robin Schiewer\textsuperscript{\rm 1},\\
Amit Kumar Pal\textsuperscript{\rm 1,2},
Anand Subramoney\textsuperscript{\rm 4},
David Kappel\textsuperscript{\rm 1}
}

\affiliations{
\small
\textsuperscript{\rm 1} Bielefeld University, Germany\\
\texttt{\{neeraj.sushma, robin.schiewer, amit.pal, david.kappel\}@uni-bielefeld.de} \\[1ex]
\textsuperscript{\rm 2} Ruhr Universit\"{a}t Bochum, Germany\\[1ex]
\textsuperscript{\rm 3} BITS Pilani, Goa Campus\\[1ex]
\textsuperscript{\rm 4} Royal Holloway, University of London, UK\\

\texttt{anand.subramoney@rhul.ac.uk} \\
}

\begin{document}

\maketitle

\begin{abstract}
Modern deep neural networks are trained using error backpropagation, which requires sequential forward and backward computations across network layers.
As these networks become deeper, this introduces limitations, since layer-wise updates are strictly interdependent and cannot proceed in parallel. These constraints restrict training procedures to data-parallel schemes, thereby prohibiting model-parallel training.
We propose \emph{TreeProp}, an architecture-agnostic variational learning framework that organizes network layers into a tree-structured hierarchy.
During training, TreeProp replaces sequential forward computations and backward gradient propagation with hierarchical computations. This allows intermediate representations and learning signals to be constructed in time complexity of \(\mathcal{O}(\log N)\) for a network of \(N\) layers.
To the best of our knowledge, TreeProp is the first learning algorithm for deep neural networks with logarithmic parallel time complexity for both forward computation and backward gradient propagation during training.
Furthermore, we show that multiple valid paths through the hierarchy exist, such that TreeProp implicitly learns subnetworks with different effective depths, 
but without additional training effort.
We evaluate TreeProp on vision classification and autoregressive language
modeling, matching the performance of conventional end-to-end training for a variety of tasks and outperforming previous contrastive training approaches. We further demonstrate the applicability of TreeProp to recurrent neural networks that otherwise rely on backpropagation through time.
\end{abstract}

\section{Introduction}

\begin{figure}[ht!]
    \centering
    \includegraphics[width=1\linewidth]{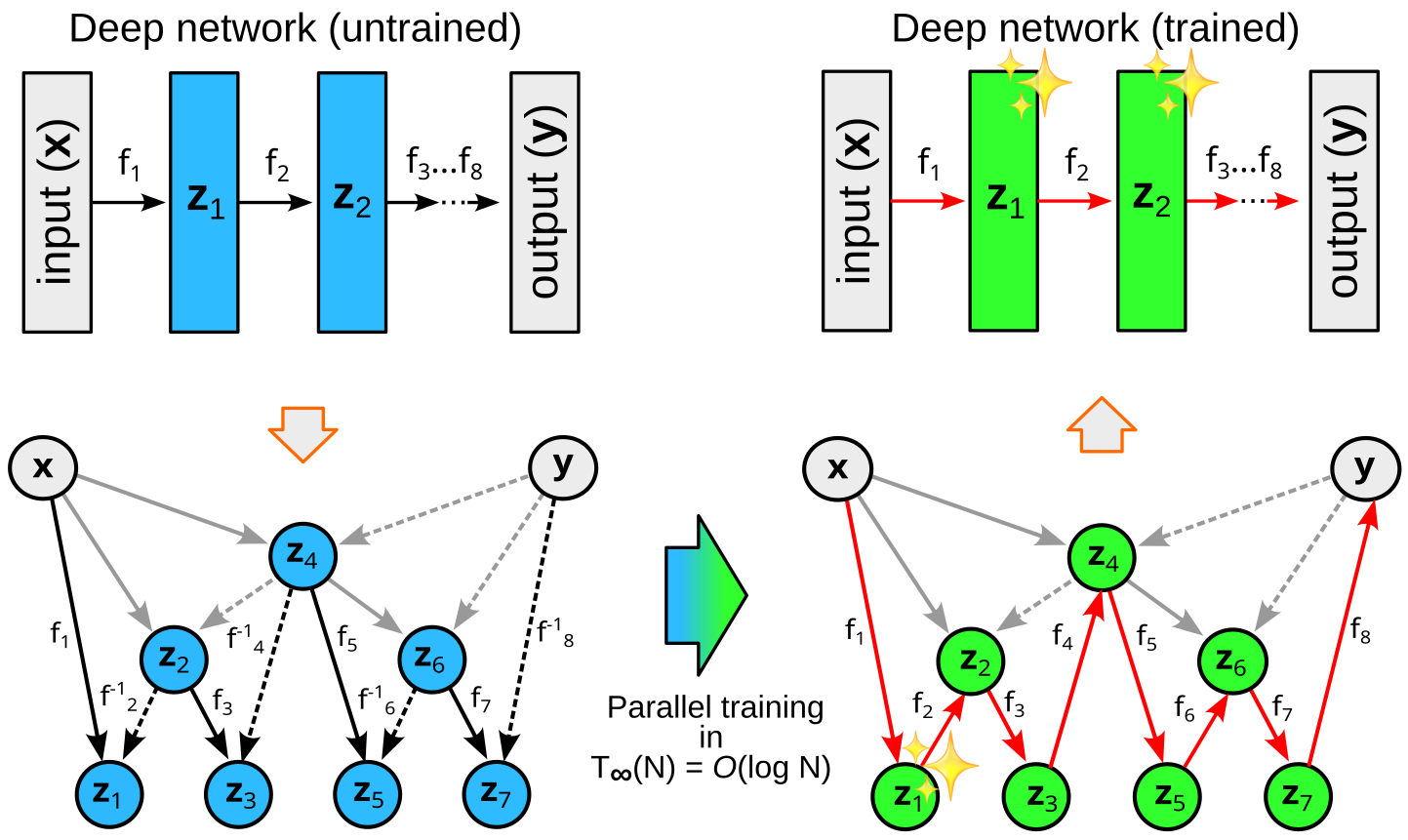}
    \caption{TreeProp reorganizes an \(N\)-layer network into a hierarchical training graph with \(T_{\infty}(N)=\mathcal{O}(\log N)\) parallel time. 
    TreeProp enables parallel training of all nodes in a given depth and
    provides the trained parameters for inference along the original sequential
    path, shown in red.
    }
    \label{fig:main-idea}
\end{figure}
\begin{figure*}[t]
    \centering
    \includegraphics[width=\linewidth]{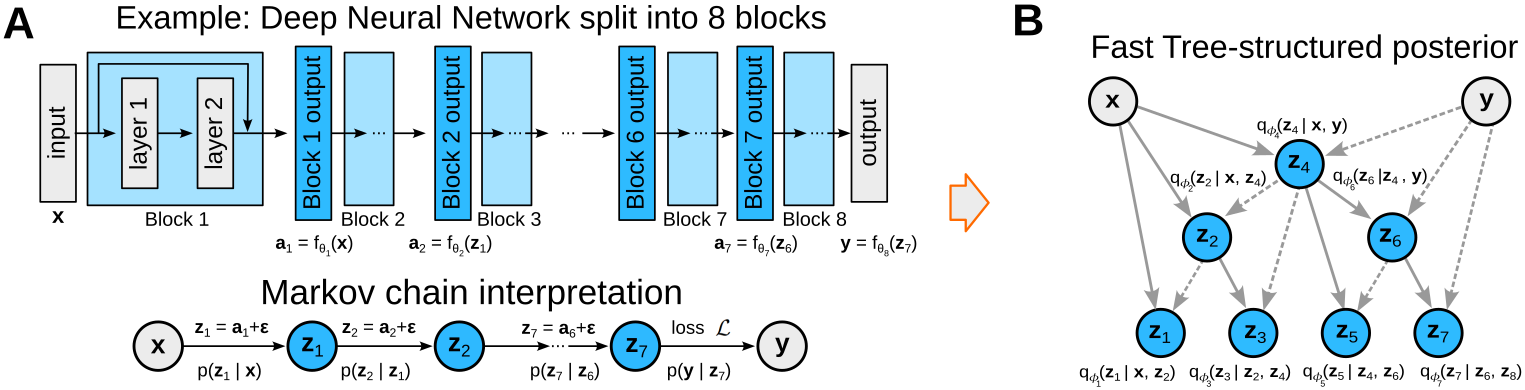}   
    \caption{\textbf{A:} A deep network and corresponding first order Markov chain interpretation. Bayesian network showing the conditional dependence properties of the model (exemplary for a network split into 8 blocks). \textbf{B:} The tree-structured variational posterior \(q_\phi(\bz_{1:N-1}\mid\bx,\by)\). Each factor infers an intermediate representation from the boundary representations of its interval, allowing all representations at the same tree level to be evaluated in parallel.}
    \label{fig:bayesian-model-illustration}
\end{figure*}
Error Backpropagation (BP) \citep{10.1007/BFb0006203,rumelhart_learning_1986} has emerged as the de facto standard method for training deep neural networks (DNNs). This method computes gradients by propagating activations sequentially through all layers of a DNN, computing the loss at the output and  propagating the gradients sequentially backward through the network. BP has demonstrated unprecedented scaling properties enabling  DNNs with trillions of parameters and close to human-level performance on complex tasks including natural language modeling. However, these gains come at increasing computational and energy costs, driven not just by the model size but also by the training paradigm itself.

One major disadvantage of BP stems from the sequential error propagation scheme. Parameter updates are locked until a full forward-backward pass across all layers of the network is performed. This locking problem limits the parallelization properties in learning and leads to reduced hardware utilization \citep{jaderberg_decoupled_2017}.
The limitations of BP have motivated the search for alternative learning methods that relax or remove the need for global error propagation. In particular, local learning approaches aim to train different parts of networks using locally defined rules and reducing inter-layer dependencies 
(\citealp{dfa, pmlr-v162-dellaferrera22a}). However, while such approaches have shown promise in vision and smaller-scale settings, 
most retain the original sequential forward path and therefore have \(\mathcal{O}(N)\) forward parallel time complexity for a network of \(N\) blocks, or resort to fully local approximate learning.

In this work, we propose TreeProp, a framework that enables gradient estimation in DNNs without full end-to-end backpropagation, while maintaining a well-defined global loss.
TreeProp decomposes networks into hierarchically organized blocks trained using local objectives, providing a general and theoretically grounded framework for efficient parallel learning (see Fig.~\ref{fig:main-idea}).
Based on a variational formulation of deep learning, we provide a principled probabilistic framework of learning without global backpropagation, that enables distributed training. 
By enabling communication between levels of the hierarchy through invertible transformations, TreeProp allows information to propagate across the network while avoiding the full sequential gradient dependencies.
Our results demonstrate that TreeProp scales to vision classification and autoregressive language modeling, achieving competitive performance on CIFAR-10, CIFAR-100 \citep{krizhevsky2009learning} and WikiText-103 \citep{merity2017pointer}.
We also apply TreeProp to recurrent neural networks (RNNs), enabling a \(\mathcal{O}(\log N)\) parallel-time approach
for training sequential models.

The parallel compute complexity allows us to reason about the efficiency of algorithms that run on multiple parallel resources, i.e. $T_\infty$ denotes the level of parallelization an algorithm can achieve on an optimal number of parallel compute nodes (c.f. \citet{cormen2022introduction}, p.779f). For conventional BP we have $T_\infty(N) = \mathcal{O}(N)$, for a network with $N$ layers, due to locking while forward and backward propagation. We present here an alternative algorithm with $T_\infty(N) = \mathcal{O}(\log N)$. We believe TreeProp is therefore most useful for distributed and constraint scenarios such as cloud-to-edge training systems \citep{kukreja2019training}, but can have impact far beyond these applications.
The main contributions of this paper are summarized as follows:
\begin{itemize}
    \item We propose TreeProp, a learning framework that replaces the sequential forward and backward pass in conventional DNN training with tree-structured hierarchically organized computation.
    \item We derive TreeProp from a variational probabilistic formulation and show
    that both the forward construction of intermediate representations and the
    backward learning path have
    \(\mathcal{O}(\log N)\) parallel time complexity.
    \item We show first training results, demonstrating competitive performance for feedforward and recurrent networks, including classification with Vision Transformer, and autoregressive
    language modeling.
    \item We show that TreeProp jointly learns subnetworks corresponding to
    different hierarchical paths and extend the framework to recurrent
    sequence modeling through TreeProp Through Time (TreeProp-TT).
\end{itemize}

\section{Related Work}
\label{related_works}

A wide variety of training algorithms have been introduced to replace or improve BP with the goal to improve efficiency, enable model-parallel processing or to more closely match biological learning mechanisms (or a mix of these goals).
Feedback-alignment methods use fixed or learned feedback pathways
\citep{lillicrap2014random,dfa,Kollen-Pollack,weights_mirror}, while target-propagation methods replace backpropagated errors with approximate
activation targets
\citep{lee2015difference,meulemans2020theoretical,
ernoult2022towards,shibuya2023fixed}. 
These methods relax or avoid locking during gradient propagation, but retain a sequential computation in the forward path.

Local-learning methods optimize layers or blocks using auxiliary objectives.
These include contrastive and Forward-Forward approaches
\citep{xiong_loco_2020,illing_local_2021,hinton2022forward,
zhao2023cascaded,constrative_fwd_fwd}, forward-only zeroth-order optimization \citep{malladi2023fine}, synthetic-gradient methods \citep{jaderberg_decoupled_2017}, and augmented auxiliary networks \citep{ma_scaling_2024}. Although these approaches reduce global backward
dependencies, they generally retain a sequential forward path through the network.
Layer-parallel methods exploit the residual structure of ResNets and
Transformers to replace layer-serial computation with iterative
parallel-in-time or multigrid solvers \citep{jiang2026layerparallel}. TreeProp instead applies to general sequentially composed network blocks by organizing intermediate representations into a jointly trained tree-structured hierarchy.

Probabilistic formulations provide another basis for local learning.
Probabilistic block-local learning derives local objectives from latent representations but retains an \(\mathcal{O}(N)\) forward path \citep{kappel_block-local_2023}. 
Predictive-coding methods use local prediction errors and iterative inference dynamics \citep{salvatori2025survey}. 
NoProp \citep{li2025noproptrainingneuralnetworks} and DiffusionBlocks \citep{shing2026diffusionblocks} propose distinct diffusion-based approaches and train blocks using local denoising or score-matching objectives, avoiding full backward propagation although forward propagation still retains sequential dependencies between blocks.

To the best of our knowledge, TreeProp is the first learning framework in which both the forward construction of intermediate representations and the backward learning path have \(\mathcal{O}(\log N)\) parallel time
complexity for a general DNN of \(N\) blocks. 
Existing approaches either retain sequential computation in the forward or both paths, rely on iterative inference or parallel solvers, or locally apply greedy learning.

\section{TreeProp}
We introduce TreeProp as a variational framework that reorganizes training in deep neural networks from a sequential forward-backward path of length \(\mathcal{O}(N)\) into a tree-structured path of length \(\mathcal{O}(\log N)\).
We first view a standard feedforward network as a sequential chain of intermediate representations and discuss the resulting forward-backward learning path. 
We then introduce a probabilistic interpretation in which intermediate representations are treated as latent variables conditioned on the input and target. 
This posterior view allows the sequential chain to be reorganized into a tree-structured factorization, which forms the basis for the variational objectives and logarithmic training paths developed in the following subsections.
\subsection{Neural nets, chains and trees}
Deep neural networks (DNNs) are universal function approximators that map an input $\bx$ to an output $\by = f(\bx)$, via a transfer function $f: \bx \rightarrow \by$, $(\bx, \by) \in \mathcal{D}_\bx \times \mathcal{D}_\by$, where $\mathcal{D}_\bx$, $ \mathcal{D}_\by$ are some (possibly high-dimensional) domains. Typically $f$ is constructed by chaining together a sequence of $N$ nested operations, $(f_1 \circ f_2 \circ \dots \circ f_N)(\bx) = f(\bx) = \by$, where $\ba_n = f_n(\ba_{n-1})$ are the transfer functions with activations $\ba_n$. Intuitively, $N$ can denote the number of layers in the network. However, in many modern neural network architectures it is natural to think of this chain of operations in terms of more complex functional blocks, such as transformer blocks that consist of self-attention, MLP layers, normalization and residual connections \citep{NIPS2017_3f5ee243}. The theoretical framework we present here is agnostic to the level of granularity that is chosen.
This chain structure induces a sequential computational path. In the forward pass, representations must be computed in order: 
\begin{equation*}
\bx \rightarrow \ba_1 \rightarrow \cdots \rightarrow \ba_{N-1} \rightarrow \by.
\label{eqn:deterministic-chain}
\end{equation*}
Here \(\by\) denotes the network prediction.
During training, \(\by\) is compared with the observed target
through a loss function \(\mathcal{L}\). In the backward pass,
backpropagation computes adjoints through the Jacobian-vector recursion for $n=N,\dots,1$
\begin{equation*}
\boldsymbol{\lambda}_n
:=
\nabla_{\ba_n}\mathcal{L},
\quad
J_n
:=
\frac{\partial \ba_n}{\partial \ba_{n-1}},\quad
\boldsymbol{\lambda}_{n-1}
=
J_n^\top \boldsymbol{\lambda}_n\;.
\label{eqn:backward-recursion}
\end{equation*}
Thus, each adjoint depends on the next adjoint in the chain, yielding the
backward sequential dependency
\begin{equation}
\boldsymbol{\lambda}_{N-1}
\rightarrow
\boldsymbol{\lambda}_{N-2}
\rightarrow
\cdots
\rightarrow
\boldsymbol{\lambda}_{1}
\rightarrow
\boldsymbol{\lambda}_{0}.
\label{eqn:backward-chain}
\end{equation}
Both the forward computation of representations and the backward propagation of
adjoints therefore have critical-path length \(\mathcal{O}(N)\).

To study these chain structures in more depth, it can be insightful to adopt a probabilistic interpretation of the operations performed by a DNN.
We assume that training examples are drawn from an unknown data distribution \(p^*(\bx,\by)\).
A neural network with parameters \(\theta\) defines a conditional predictive distribution \(p_\theta(\by\mid\bx)\) through a sequence of
intermediate representations.
Training then corresponds to minimizing the expected negative log-likelihood 
\begin{equation*}
\mathcal{L}(\theta)
=
\mathbb{E}_{(\bx,\by)\sim p^*}
\left[
-\log p_\theta(\by\mid\bx)
\right].
\label{eqn:standard-nll}
\end{equation*}
This objective is empirically approximated using minibatches from the training set.
In classification, \(p_\theta(\by\mid\bx)\) is typically a categorical distribution parameterized by a softmax output, yielding the cross-entropy loss. 
In regression, a Gaussian likelihood with fixed variance yields the mean squared error loss.

To also make the intermediate computations probabilistic, we use the reparameterization trick \citep{kingma_AutoEncoding_2013} and introduce stochastic latent representations \(\bz_n = \ba_n + \beps_n\), where \(\beps_n\) is independent noise injected at the \(n\)-th intermediate representation.
Let $\{\bz_1, \dots, \bz_{N-1}\}$ be the latent representations across the blocks, then the DNN can be defined as a directed graphical model of the form of a first-order Markov chain (see Fig.~\ref{fig:bayesian-model-illustration}A for an illustration):
\begin{equation*}
\bx \rightarrow \bz_1 \rightarrow \bz_2 \rightarrow \cdots \rightarrow \bz_{N-1} \rightarrow \by.
\end{equation*}



To simplify the notation we define $\bz_0=\bx$ and $\bz_N=\by$ to arrive at a compact conditional probabilistic model over the intermediate representations and the output:
\begin{align}
p_\theta(\bz_{1:N-1},\by \mid \bx)
&=
p_{\theta_N}(\by \mid \bz_{N-1})
\prod_{n=1}^{N-1}
p_{\theta_n}(\bz_n \mid \bz_{n-1}), \nonumber \\
&=
\prod_{n=1}^{N}
p_{\theta_n}(\bz_n \mid \bz_{n-1})\;.
\label{eqn:conditional-chain}
\end{align}
Here \(p_{\theta_n}(\bz_n\mid\bz_{n-1})\) denotes the stochastic transition
implemented by the \(n\)-th network layer, and \(p_{\theta_N}(\by\mid\bz_{N-1})\) denotes the output likelihood of the observed target.

In supervised training, each example provides both an input \(\bx\) and an
observed target \(\by\). The probabilistic interpretation of DNNs allows us to explore also other training schemes than BP, such as Expectation-Maximization or variational learning \cite{dempster1977maximum, kingma_AutoEncoding_2013}. However, these methods require us to evaluate the posterior distribution over latent intermediate representations
\(\bz_{1:N-1}\), given observed inputs and targets. Thus, we seek the training-time posterior
\begin{equation}
p_\theta(\bz_1,\dots,\bz_{N-1}\mid \bx,\by)
=
\frac{
p_\theta(\bz_1,\dots,\bz_{N-1},\by\mid \bx)
}{
p_\theta(\by\mid \bx)
}\;.
\label{eqn:true-posterior}
\end{equation}
This posterior assigns high probability to intermediate representations that
explain the observed target while remaining consistent with the forward
pass through of the network.

In principle, learning could proceed by inferring the intermediate
representations from the posterior in Eq.~\eqref{eqn:true-posterior} and using
them to optimize the model parameters. In practice, this presents two
challenges. First, exact posterior inference is generally intractable because
the normalizing term
\begin{equation*}
p_\theta(\by\mid\bx)
=
\int
p_\theta(\bz_{1:N-1},\by\mid\bx)
\,d\bz_{1:N-1}
\label{eqn:marginal-predictive}
\end{equation*}
requires marginalizing over all intermediate latent representations.
Variational learning addresses this challenge by introducing a tractable approximate posterior \(q_\phi(\bz_{1:N-1}\mid\bx,\by)\), whose parameters \(\phi\) are learned to approximate the exact posterior \(p_\theta(\bz_{1:N-1}\mid\bx,\by)\).

A second challenge, which constitutes the central problem addressed in this work, is that both inference and learning inherit the sequential dependencies of the original chain. Target information must pass through successive intermediate representations to influence earlier parts of the network. 
Since \(\bz_n=\ba_n+\beps_n\), the latent representations also inherit the same reverse dependency as in Eq.~\eqref{eqn:backward-chain}. 
Consequently, a variational distribution that preserves the original chain structure would still retain
\(\mathcal{O}(N)\) forward and backward learning paths, while a fully factorized form, if not chosen carefully, would degrade performance.
We therefore seek a variational posterior that retains a close approximation to Eq.~\eqref{eqn:true-posterior} while admitting efficient parallel computation. 
To ensure that this, the next subsection first shows that the true posterior itself can be decomposed into an equivalent hierarchical tree-structured factorization. 
Building on this result, we then introduce the corresponding tree-structured variational posterior and derive its learning objective.


\subsection{Tree-Structured Posterior Factorization}
\label{sec:tree-construction}

We next describe how the sequential posterior is reorganized into a tree and introduce the notation used throughout the remainder of the paper. 
Consider first the special case \(N=2^D\),
where \(D\in\mathbb{N}\) denotes the depth of the hierarchy, with
\(\bz_0=\bx\) and \(\bz_N=\by\).

Each intermediate representation \(\bz_n\), for
\(n\in\{1,\dots,N-1\}\), is introduced at tree level
\begin{equation*}
d(n)
:=
\max\left\{
k\in\mathbb{N}_0
\,:\,
2^k \text{ divides } n
\right\}.
\end{equation*}
For a fixed node \(n\), let
\begin{equation*}
l:=n-2^{d(n)},
\qquad
r:=n+2^{d(n)}.
\end{equation*}
Then \(\bz_l\) and \(\bz_r\) are the boundary representations of the interval \([l,r]\) in which \(\bz_n\) is introduced. 
Writing \(\operatorname{pa}(n):=(\bz_l,\bz_r)\), the corresponding tree-structured posterior factor becomes \(p_\theta(\bz_n\mid\operatorname{pa}(n)) = p_\theta(\bz_n\mid\bz_l,\bz_r)\).

\begin{theorem}[Chains-to-trees posterior factorization]
\label{thm:chains-to-trees}
Let \(\bx,\bz_{1:N-1},\by\) be random variables whose conditional distribution factorizes according to the first-order Markov chain in Eq.~\eqref{eqn:conditional-chain}, and let \(N=2^D\). 
Then the posterior over the intermediate representations also admits the hierarchical factorization
\begin{equation*}
p_\theta(\bz_{1:N-1}\mid\bx,\by)
=
\prod_{n=1}^{N-1}
p_\theta\!\left(
\bz_n\mid\operatorname{pa}(n)
\right).
\label{eqn:chains-to-trees-factorization}
\end{equation*}
\end{theorem}

It follows by recursively applying the posterior identity
\begin{equation}
p_\theta(\bz_n\mid\bz_l,\bz_r)
=
\frac{
p_\theta(\bz_n\mid\bz_l)\,
p_\theta(\bz_r\mid\bz_n)
}{
p_\theta(\bz_r\mid\bz_l)
}
\label{eq:local-bridge-identity}
\end{equation}
at each node \(n\) of the hierarchy. Conditioning on the midpoint representation \(\bz_n\) that separates the left and right subchains by the first-order Markov property. 
The same decomposition can therefore be applied
recursively to the subintervals \([l,n]\) and \([n,r]\). 
A complete inductive proof of Theorem~\ref{thm:chains-to-trees} is provided in Appendix~\ref{app:proof-chains-to-trees} of the supplementary material.

For clarity, Theorem~\ref{thm:chains-to-trees} is stated for the special case \(N=2^D\). 
For arbitrary \(N\), the same construction is obtained by
recursively selecting, for each interval \([l,r]\), the intermediate index \(n=\left\lfloor (l+r)/2 \right\rfloor\),
The representations \(\bz_l\) and \(\bz_r\) again serve as the left and right boundary representations of \(\bz_n\). Recursively splitting the resulting unequal subintervals yields the same tree-structured posterior factorization,
with hierarchy depth at most \(\lceil\log_2 N\rceil\).

\subsection{Variational Learning with Tree Factorization}
\label{sec:variational-tree-learning}


\begin{figure}
    \centering
    \includegraphics[width=0.70\linewidth]
    {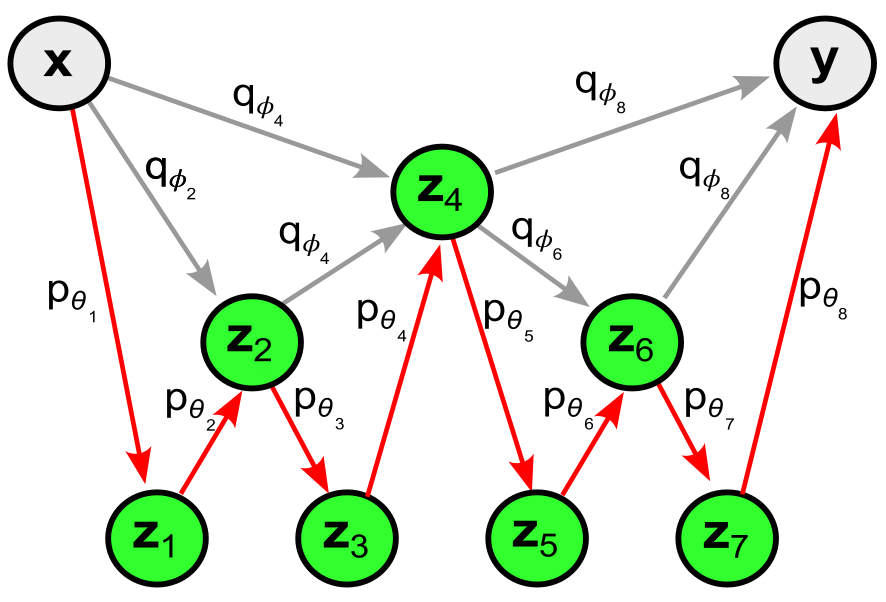}
\caption{\textbf{TreeProp after training:}
Red arrows denote the trained factors of the original sequential model, parameterized by \(p_{\theta_n}(\bz_n\mid\bz_{n-1})\).
Gray arrows denote the trained auxiliary tree-structured factors introduced
by TreeProp parameterized by \(q_{\phi_n}\). These auxiliary parameters define multiple valid subnetworks with
different effective depths, including the original sequential path and
shorter paths such as
\(\bx\rightarrow\bz_4\rightarrow\by\),
\(\bx\rightarrow\bz_4\rightarrow\bz_6\rightarrow\bz_7\rightarrow\by\) etc.}
    \label{fig:tree-prop}
\end{figure}

Theorem~\ref{thm:chains-to-trees} establishes that the exact training-time
posterior admits a tree-structured factorization. Motivated by this result,
TreeProp adapts this same dependency structure for its tractable approximate posterior:
\begin{equation}
q_\phi(\bz_{1:N-1}\mid\bx,\by)
=
\prod_{n=1}^{N-1}
q_{\phi_n}\!\left(
\bz_n\mid\operatorname{pa}(n)
\right).
\label{eqn:tree-variational-posterior}
\end{equation}
This tree-structured factorization is illustrated in Fig.~\ref{fig:bayesian-model-illustration}B.
Each tree-structured variational factor \(q_{\phi_n}(\bz_n\mid\operatorname{pa}(n))\) infers \(\bz_n\) from its boundary representations, allowing all nodes at the same tree level to be evaluated in parallel.
Applying the standard variational bound to the Markov-chain model using the
tree-structured variational posterior in
Eq.~\eqref{eqn:tree-variational-posterior} gives
\begin{equation*}
-\log p_\theta(\by\mid\bx)
\leq
\mathcal{F}(\theta,\phi),
\end{equation*}
where
\begin{equation}
\begin{aligned}
\mathcal{F}(\theta,\phi)
&=
\mathbb{E}_{q_\phi}\!
\Bigg[
\sum_{n=1}^{N-1}
\log
\frac{
q_{\phi_n}\!\left(\bz_n\mid\operatorname{pa}(n)\right)
}{
p_{\theta_n}(\bz_n\mid\bz_{n-1})
}
\\
&\qquad\qquad
-
\log p_{\theta_N}(\by\mid\bz_{N-1})
\Bigg]\;,
\end{aligned}
\label{eqn:tree-variational-objective}
\end{equation}
and where, \(\theta\) denotes the parameters
of the original sequential model, whereas \(\phi\) parameterizes the
tree-structured variational posterior. 
Thus, the global likelihood objective decomposes into node level loss terms associated with the constructed tree. 

A complete derivation of Eq.~\eqref{eqn:tree-variational-objective} is provided in Appendix~\ref{app:proof-variational-objective}, and further details to node-wise losses in Appendix ~\ref{app:nodewise-losses}.
Each term of Eq.~\eqref{eqn:tree-variational-objective} compares \(q_{\phi_n}(\bz_n\mid\operatorname{pa}(n))\)
with the corresponding forward transition
\(p_{\theta_n}(\bz_n\mid\bz_{n-1})\).
The final term evaluates the likelihood of the observed target.
 Importantly, this loss is constructed only using samples $\bz_n$ from the tree structured variational posterior \eqref{eqn:tree-variational-posterior}, such that no full forward pass through the DNN has to be performed at any point.
 In Appendix~\ref{app:TreeProp-gradient-bias} we derive the bias of the gradient of the TreeProp objective relative to the original end-to-end gradient.
 For this to work, the model introduces additional network blocks with new parameters that function as `short-cuts' through the architecture in higher levels of the tree. The parameters $\phi$ therefore contain original DNN parameters $\theta$ and combine them with new ones. Fig.~\ref{fig:tree-prop} illustrates the parameterization of the tree model.

\subsection{Tree-Structured Variational Factor Estimation}
\label{sec:local-posterior-inference}

\begin{figure}
    \centering
    \includegraphics[width=0.8\linewidth]
    {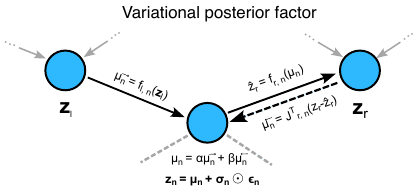}
\caption{\textbf{Estimation of Tree-structured variational factors.}
The left boundary \(\bz_l\) produces the forward estimate
\(\boldsymbol{\mu}_n^{\rightarrow}\), while the right-boundary prediction
error produces the correction \(\boldsymbol{\mu}_n^{\leftarrow}\).
The coefficients \(\alpha,\beta\in[0,1]\) control the relative contributions
of the forward estimate and backward correction, respectively. Their weighted
combination,
\(\boldsymbol{\mu}_n=
\alpha\boldsymbol{\mu}_n^{\rightarrow}
+\beta\boldsymbol{\mu}_n^{\leftarrow}\),
parameterizes \(q_{\phi_n}(\bz_n\mid\bz_l,\bz_r)\), from which
\(\bz_n\) is sampled.}
    \label{fig:variational-posterior}
\end{figure}

For each interval \([l,r]\), TreeProp infers an intermediate
representation \(\bz_n\) conditioned on the boundary
representations \(\bz_l\) and \(\bz_r\), as illustrated in
Fig.~\ref{fig:variational-posterior}.  
To implement these posteriors we use normalizing flows \cite{rezende2015variational}.
Let \(f_{l,n}\) and \(f_{r,n}\) denote the
neural network block associated with the left and right parts of the interval. 
At the lowest tree level, \(f_{l,n}=f_n\) and \(f_{r,n}=f_{n+1}\); at higher tree levels, these are
additional blocks parameterized by auxiliary parameters in \(\phi\).

The left block produces the forward estimate
\begin{equation*}
\boldsymbol{\mu}_n^{\rightarrow}
=
f_{l,n}(\bz_l),
\end{equation*}
which is propagated through the right block to predict the right boundary,
\begin{equation*}
\widehat{\bz}_r
=
f_{r,n}
\bigl(\boldsymbol{\mu}_n^{\rightarrow}\bigr).
\end{equation*}

We estimate the backward propagation of right-boundary using the Jacobian
\begin{equation*}
\mathbf{J}_{r,n}
=
\left.
\frac{\partial f_{r,n}(\bz)}
     {\partial \bz}
\right|_{\bz=\boldsymbol{\mu}_n^{\rightarrow}} .
\end{equation*}
The right-boundary prediction error defines the correction
\begin{equation*}
\boldsymbol{\mu}_n^{\leftarrow}
=
\mathbf{J}_{r,n}^{\top}
\bigl(
\bz_r-\widehat{\bz}_r
\bigr).
\end{equation*}
The variational posterior mean is
\begin{equation}
\boldsymbol{\mu}_n
=
\alpha\boldsymbol{\mu}_n^{\rightarrow}
+
\beta\boldsymbol{\mu}_n^{\leftarrow},
\label{eqn:posterior-mean}
\end{equation}
where \(\alpha,\beta\in[0,1]\) are fixed hyperparameters controlling the forward estimate and right-boundary correction, respectively.
The tree-structured variational factor is centered at this posterior mean,
and \(\bz_n\) is sampled from the resulting distribution. In the Gaussian
case,
\begin{equation}
\bz_n
=
\boldsymbol{\mu}_n
+
\boldsymbol{\sigma}_n
\odot
\boldsymbol{\epsilon}_n,
\qquad
\boldsymbol{\epsilon}_n
\sim
\mathcal{N}(\mathbf{0},\mathbf{I}).
\label{eqn:midpoint-reparameterization}
\end{equation}

The complete inference procedure draws a reparameterized sample from
\(q_{\phi_n}(\bz_n\mid\bz_l,\bz_r)\).

Applying this inference procedure recursively constructs the hierarchy shown in Fig.~\ref{fig:bayesian-model-illustration}B. 
We next analyze the sequential depth of the resulting forward and backward computations.

\subsection{Logarithmic Forward-Backward Learning Path}
\label{sec:logarithmic-learning-path}

In the preceding subsections, we derived the tree-structured variational
objective in Eq.~\eqref{eqn:tree-variational-objective} and introduced the
corresponding inference procedure in Eq.~\eqref{eqn:midpoint-reparameterization},
which constructs the hierarchical representations. 
We now show that, whereas representation construction and gradient
propagation each have sequential depth \(\mathcal{O}(N)\) in a standard
\(N\)-block network, TreeProp evaluates nodes at the same hierarchy depth
in parallel, reducing the forward-backward learning path to
\(\mathcal{O}(\log N)\).

\begin{theorem}[Logarithmic forward-backward learning path]
\label{thm:logarithmic-learning-path}
Let every nontrivial interval \([l,r]\) introduce the node
\(n=\left\lfloor(l+r)/2\right\rfloor\) by estimating the tree-structured
variational factor \(q_{\phi_n}(\bz_n\mid\bz_l,\bz_r)\) and drawing a
reparameterized sample from it.
Assume that gradients are propagated through the resulting computation graph, with adjoints
\[
\boldsymbol{\lambda}_n
:=
\nabla_{\bz_n}\mathcal{F}
\in\mathbb{R}^{\dim(\bz_n)}.
\]
Then the forward computation of all intermediate representations and the
backward computation of all adjoints each have critical-path depth \(D=\lceil\log_2 N\rceil\).
Consequently, the complete forward-backward learning path has
\(\mathcal{O}(\log N)\) parallel time complexity. 

\end{theorem}

A complete proof of Theorem~\ref{thm:logarithmic-learning-path} is provided in Appendix~\ref{app:proof-logarithmic-learning-path}.

Algorithm~\ref{alg:TreeProp-training} implements this construction.
At each tree level, all intervals whose boundary representations are
available are processed in parallel. For each interval \([l,r]\), the
corresponding midpoint representation is obtained by drawing a reparameterized sample from the
tree-structured variational factor \(q_{\phi_n}(\bz_n\mid\bz_l,\bz_r)\). 
The sampled representation \(\bz_n\) then becomes a boundary for
the two child intervals \([l,n]\) and \([n,r]\).

For each such interval, let
\begin{equation}
\mathcal{F}_{l,n,r}
=
\mathcal{J}_{l,n,r}
\left(
\bz_l,\bz_n,\bz_r;\theta,\phi
\right)
\label{eq:interval-local-objective}
\end{equation}
denote its contribution to the training objective. At the lowest tree
level, \(r-l=2\), so that \(l=n-1\) and \(r=n+1\). In this case,
\begin{equation}
\mathcal{F}_{n-1,n,n+1}
=
\log
\frac{
q_{\phi_n}
\left(
\bz_n\mid\bz_{n-1},\bz_{n+1}
\right)
}{
p_{\theta_n}
\left(
\bz_n\mid\bz_{n-1}
\right)
},
\label{eq:finest-level-local-objective}
\end{equation}
which is the corresponding node-wise transition term in
Eq.~\eqref{eqn:tree-variational-objective}. Thus, the original
adjacent-chain factor \(p_{\theta_n}(\bz_n\mid\bz_{n-1})\) appears directly at the lowest tree level.

The resulting objective \(\mathcal{F}_{\mathrm{tree}}\) is differentiated
through the tree-structured computation graph rather than through the
original sequential computation graph.
The resulting tree-induced paths, path-specific training objectives, and parameter overhead are discussed in Appendix~\ref{app:subnetwork_analysis}.
The additional neural network blocks in higher levels of the tree introduce additional parameters that are used during training to compute short-cut paths. 
Figure ~\ref{fig:tree-prop} illustrates how these additional parameters implicitly induces subnetworks with different effective depths, e.g. the path $\bx \rightarrow \bz_4 \rightarrow \bz_6 \rightarrow \bz_7 \rightarrow \by$. 
In experiments in Appendix~\ref{app:subnetwork_analysis},
we show that these networks emerge as functional shallow network through the same learning procedure.

\begin{algorithm}[tb]
\caption{TreeProp}
\label{alg:TreeProp-training}
\textbf{Input}: minibatch \((\bx,\by)\), network with \(N\) blocks,
hierarchy depth \(D=\lceil\log_2 N\rceil\), tree-structured variational factors \(\{q_{\phi_n}(\bz_n\mid\operatorname{pa}(n))\}_{n=1}^{N-1}\),
node objectives \(\{\mathcal{F}_{l,n,r}\}\), and noise scales
\(\{\sigma_n\}_{n=1}^{N-1}\)
\par\medskip
\textbf{Forward tree propagation}
\begin{algorithmic}[1]
\STATE \(\bz_0\leftarrow\bx,\quad\bz_N\leftarrow\by\),\quad
\(
\mathcal{B}_1\leftarrow\{(0,N)\}
\)
\FOR{\(s=1,2,\ldots,D\)}
    \FORALL{\([l,r]\in\mathcal{B}_s\) \hl{in parallel}}
        \IF{\(r-l\geq2\)}
            \STATE
            \(
            \bz_n
            \sim
            q_{\phi_n}(\bz_n\mid\bz_l,\bz_r)
            \)
            \STATE \(\mathcal{F}_{l,n,r}\leftarrow
            \log
            \frac{
            q_{\phi_n}(\bz_n\mid\bz_l,\bz_r)
            }{
            p_{\theta_n}(\bz_n\mid\bz_{n-1})
            }\)
            \IF{\(r=N\ \AND\ r-l=2\)}
                \STATE \(\mathcal{F}_{l,n,r}\leftarrow
                \mathcal{F}_{l,n,r}
                -\log p_{\theta_N}(\by\mid\bz_{N-1})\)
            \ENDIF
            \STATE \(\mathcal{B}_{s+1}\leftarrow
            \mathcal{B}_{s+1}\cup\{(l,n),(n,r)\}\)
        \ENDIF
    \ENDFOR
\ENDFOR
\end{algorithmic}
\par\medskip
\textbf{Backward tree propagation}
\begin{algorithmic}[1]
\STATE
\(\boldsymbol{g}_{\theta,\phi}
\leftarrow\boldsymbol{0}\)
\FOR{\(s=D,D-1,\ldots,1\)}
    \FORALL{\([l,r]\in\mathcal{B}_s\) \hl{in parallel}}

        \STATE
        \(
        \boldsymbol{g}_{\theta,\phi}
        \leftarrow
        \boldsymbol{g}_{\theta,\phi} +
        \nabla_{\theta,\phi}
        \mathcal{F}_{l,n,r}
        \)
        
    \ENDFOR
\ENDFOR

\STATE \hl{In parallel:} Update \(\theta\) and \(\phi\) using 
\(\boldsymbol{g}_{\theta,\phi}\)


\end{algorithmic}
\end{algorithm}

\section{Experiments}
\label{sec:experiments}
We evaluate TreeProp on feedforward classification, Vision Transformer classification, autoregressive language modeling, and recurrent sequence modeling. 
Across experiments, we compare primarily against standard
end-to-end backpropagation (BP) using matched architectures and training settings. 
Where applicable, we also report contrastive Forward-Forward
(CFF+M) results \citep{constrative_fwd_fwd} since it has demonstrated that Forward-Forward style training can scale to modern architectures such as Vision Transformers and therefore provides a relevant alternative baseline to backpropagation.
Unless stated otherwise, all results compare models with matched inference-time parameter counts.
Implementation details and additional analyses are provided in
Appendix~\ref{app:experimental-details}.
We additionally ablate the contributions of the node-level KL and
intermediate-level cross-entropy losses across TreeProp's induced subnetworks, as well as the weighting of the forward estimate and right-boundary correction through \((\alpha,\beta)\) in Appendix~\ref{app:ablation}.

\subsection{MNIST Classification}

First, to demonstrate the concept, we evaluate TreeProp on MNIST \citep{726791} using sigmoid MLPs without
residual connections or normalization. We compare against BP and the reported CFF+M baseline across
hierarchy depths \(2,3,\) and \(4\), corresponding to networks with
\(4,8,\) and \(16\) layers. Table~\ref{tab:classification_results} reports
the depth-\(3\) setting with an 8-layer MLP and hidden size \(500\). 
TreeProp remains close to BP while outperforming the reported CFF+M baseline. 
Complete results across hierarchy depths and hidden sizes are provided in Table~\ref{tab:mnist_full_result} 
in Appendix~\ref{app:mnist_details}.
See Appendix~\ref{app:mnist_details} for more details.

\subsection{Vision Transformer Classification}

We evaluate TreeProp on CIFAR-10, CIFAR-100, and TinyImageNet using
Vision Transformer-style architectures \citep{dosovitskiy2021an}.
For TreeProp, the Transformer blocks are organized into a depth-\(3\)
hierarchy. We compare against separately trained end-to-end BP models with
matched effective depth and the CFF+M baseline
\citep{constrative_fwd_fwd}. Table~\ref{tab:classification_results}
summarizes the top-1 classification results. TreeProp remains competitive with BP across all three datasets, closely
matching it on CIFAR-10 and outperforming it on CIFAR-100 and TinyImageNet.
It also consistently outperforms the reported CFF+M baseline.
See Appendices~\ref{app:cifar10_details},
\ref{app:cifar100_details}, and
\ref{app:tinyimagenet_details} for further details.

\begin{table*}[t]
\centering
\small
\begin{tabular}{lcc|cc|cc|cc|c}
\toprule
& \multicolumn{2}{c|}{MNIST}
& \multicolumn{2}{c|}{CIFAR-10}
& \multicolumn{2}{c|}{CIFAR-100}
& \multicolumn{2}{c|}{TinyImageNet}
& \begin{tabular}{c}
Ideal \(T_\infty\)\\
speedup
\end{tabular} \\
Method
& Acc. & Params.
& Acc. & Params.
& Acc. & Params.
& Acc. & Params.  \\
\\
\midrule
BP
& $98.51 \pm 0.05$ & 1.89M
& $88.62 \pm 0.84$ & 3.27M
& $59.12 \pm 0.48$ & 3.30M
& $43.91 \pm 0.15$ & 3.35M
& $1\times$ \\

CFF+M
& $96.12 \pm 0.05$ & 1.89M
& $83.87 \pm 0.65$ & 3.27M
& $58.60 \pm 0.72$  & 3.30M
& $40.92 \pm 0.63$ & 3.35M
& $1.78\times$  \\

TreeProp
& $98.38 \pm 0.13$ & 1.89M
& $88.71 \pm 0.35$ & 3.27M
& $62.69 \pm 0.49$ & 3.30M
& $45.23 \pm 0.20$ & 3.35M
& $2.67\times$ \\
\bottomrule
\end{tabular}
\caption{Classification results and parameter counts for TreeProp and
corresponding baselines. TreeProp is trained using a depth-\(3\) hierarchy,
while BP and CFF+M use equivalent 8-layer architectures. The MNIST models
use sigmoid MLPs with hidden size \(500\), while the remaining models use
Vision Transformer architectures. BP denotes separately trained end-to-end
baselines, and CFF+M denotes the contrastive Forward-Forward baseline.The ideal \(T_\infty\) speedup is measured relative to BP by comparing the combined forward-backward critical-path depths, assuming sufficient parallel resources and equal unit cost per computational step. Higher accuracy and higher \(T_\infty\) speedup are better.}
\label{tab:classification_results}
\end{table*}

\subsection{Autoregressive Language Modeling}

We evaluate TreeProp on autoregressive language modeling using WikiText-103 \citep{merity2017pointer}, using GPT-2-style architectures
\citep{karpathy2022mingpt}, where the model predicts the next token and performance is measured using perplexity (PPL). 
Each TreeProp layer corresponds to one conventional Transformer block. Additional training details are
provided in Appendix~\ref{app:language_modeling_details}.
\begin{table}[t]
\centering
\small
\setlength{\tabcolsep}{3pt}
\begin{tabular}{cc|cc|c|c}
\toprule
\multicolumn{2}{c|}{Model}
& \multicolumn{2}{c|}{Perplexity}
& Params.
& \begin{tabular}{c}
Ideal \(T_\infty\)\\speedup
\end{tabular} \\
\cmidrule(lr){1-2}
\cmidrule(lr){3-4}
D
& N
& BP
& TreeProp
& \\
\midrule

4 & 12
& $16.42 \pm 0.02$
& $18.85 \pm 0.11$
& 124.44M
& $3.25\times$ \\

5 & 24
& $16.03 \pm 0.07$
& $18.19 \pm 0.08$
& 354.83M
& $5.00\times$ \\

6 & 36
& $15.83 \pm 0.05$
& $17.87 \pm 0.12$
& 774.03M
& $6.17\times$ \\
\bottomrule
\end{tabular}
\caption{Language-modeling perplexity where lower is better, parameter counts correspond to the models retained after training, and ideal critical-path speedup relative to BP.}
\label{tab:wikitext-main-comparison}
\end{table}
Table~\ref{tab:wikitext-main-comparison} shows that TreeProp remains competitive with matched BP baselines at depths \(4\), \(5\), and \(6\), corresponding to the GPT-2 Small, Medium, and Large models. TreeProp could scale to deep LLMs with up to 36 layers without end-to-end training. While not fully matching performance, TreeProp shows similar scaling properties to BP.

\subsection{Recurrent Sequence Modeling}

TreeProp generalizes to recurrent sequence modeling.
Recurrent hidden-state transitions are Markovian and therefore compatible
with the factorization in
Theorem~\ref{thm:chains-to-trees}. 
We organize the hierarchy over time, with a node at depth \(d(n)\) computing
\begin{equation}
\bh_n =
f_{d(n)}
\left(
\bh_{n-2^{d(n)}},
\bx_{n-2^{d(n)}:n-1}
\right).
\end{equation}
Only the left parent is propagated into the recurrent state, while the right parent provides the training target. 
At inference, the deepest depth (depth-0) cell is deployed as a standard recurrent model,
\(
\bh_t=f_0(\bh_{t-1},\bx_t)
\).
The complete recurrent formulation is provided in Appendix~\ref{app:recurrent_TreeProp}.

We evaluate TreeProp-TT on sequential MNIST and permuted-sequential MNIST
using GRU \citep{cho-etal-2014-learning} and gateless Elman cells
\citep{ELMAN1990179}. We compare against parameter-matched BPTT baselines
and evaluate various hierarchy depths for GRUs and Elman
cells. Full training details and depth-wise results are provided in
Appendix~\ref{app:smnist_details}.


\begin{table}[t]
\centering
\small
\setlength{\tabcolsep}{4pt}
\begin{tabular}{llc@{\hspace{12pt}}cc}
\toprule
Dataset & Cell & Params. & BPTT & TreeProp \\
\midrule
\multicolumn{5}{@{}l@{}}{\(D{=}6,\ N{=}64\)\hfill ideal \(T_\infty\) speedup \(= 9.8\times\)} \\
\cmidrule(lr){1-5}
MNIST   & GRU   & 12.70M & 98.67 $\pm$ 0.06  & 98.07 $\pm$ 0.11 \\
MNIST   & Elman & 4.24M  & 69.03 $\pm$ 49.74 & 97.77 $\pm$ 0.09 \\
psMNIST & GRU   & 12.70M & 96.57 $\pm$ 0.23  & 97.50 $\pm$ 0.22 \\
psMNIST & Elman & 4.24M  & 96.30 $\pm$ 0.48  & 96.95 $\pm$ 0.15 \\
\addlinespace
\multicolumn{5}{@{}l@{}}{\(D{=}8,\ N{=}256\)\hfill ideal \(T_\infty\) speedup \(= 30.1\times\)} \\
\cmidrule(lr){1-5}
MNIST   & GRU   & 12.64M & 98.47 $\pm$ 0.09  & 97.53 $\pm$ 0.11 \\
MNIST   & Elman & 4.23M  & 11.35 $\pm$ 0.00  & 92.26 $\pm$ 0.45 \\
psMNIST & GRU   & 12.64M & 94.61 $\pm$ 0.14  & 95.08 $\pm$ 1.29 \\
\bottomrule
\end{tabular}
\caption{Recurrent classification on MNIST and psMNIST, using test accuracy at best validation epoch. TreeProp denotes the deployed recurrent cell and BPTT a matched
recurrent baseline. For all runs, the longest (length $D$) tree readout accuracy was greater than 98.5\,\%.}
\label{tab:recurrent_results}
\end{table}

\subsubsection{Results}

As shown in Table~\ref{tab:recurrent_results},
TreeProp-TT is competitive with matched BPTT baselines across tasks and cell types. On psMNIST, TreeProp uniformly performs better than BPTT, and on MNIST it is comparable for GRU cells, and beats BPTT for Elman cells. Elman-BPTT entries at
$11.4$ correspond to majority-class collapse (chance); in the $D{=}6$
MNIST Elman-BPTT one of three seeds collapsed. As a positive side-effect, TreeProp helps to avoid collapse due to vanishing or exploding gradients. For more results refer to Table~\ref{tab:TreeProp-all} in Appendix~\ref{app:recurrent_TreeProp}.

\section{Conclusion}

In this paper we introduce TreeProp, a tree-structured variational training framework that relaxes full end-to-end gradient propagation through DNNs.
TreeProp achieves competitive performance across feedforward classification, Vision Transformer classification, and recurrent sequence modeling. We also show for the first time complex  autoregressive language modeling results with up to 36 layers without end-to-end training.
The hierarchical factorization reduces the forward and backward learning paths to \(\mathcal{O}(\log N)\) parallel time, while also jointly learning subnetworks.
We demonstrate that additional parameters introduced by the method can be discarded after training or utilized as functional subnetworks.
Future work will investigate the roles of optimization and approximate posterior, and under which conditions baseline performance can be fully matched.
Furthermore we will focus on scaling TreeProp to larger models and distributed
settings, improving its variational objectives and approximate inversion
mechanisms, and studying probabilistic properties such as uncertainty
estimation and robustness. 
We will also further explore the learned subnetworks for adaptive inference, including speculative decoding.


\subsection*{Acknowledgments}

NMS was funded by the German Federal Ministry of Research, Technology and Space (BMFTR) project SMILES (01GQ2504A). RS and AKP were funded by BMFTR project ESCADE (01MN23004D).
The authors gratefully acknowledge the Gauss Centre for Supercomputing e.V. (\url{www.gauss-centre.eu} for funding this project by providing computing time on the GCS Supercomputer JUWELS at J{\"u}lich Supercomputing Centre (JSC).

\subsection*{Author contributions}

NMS, AN, DK, CTF, AKP and RS performed the experiments. DK, NMS and AN developed the theory. NMS, DK, AN, CTF, RS, AKP and AS wrote the paper.

\clearpage
\appendix
\setcounter{secnumdepth}{2}


\section{ Proofs and Additional Theoretical Details}

This section provides detailed proofs of the theoretical results underlying
TreeProp. 
We first prove that the posterior of the original Markov chain admits
the tree-structured factorization stated in Theorem~\ref{thm:chains-to-trees}. 
We then derive the corresponding variational upper bound in Theorem~\ref{thm:variational-tree-objective}. 
Finally, we prove the \(\mathcal{O}(\log N)\) parallel time complexity of the hierarchical forward and backward learning paths established in Theorem~\ref{thm:logarithmic-learning-path}.

\subsection{Proof of Theorem~\ref{thm:chains-to-trees}}
\label{app:proof-chains-to-trees}

We prove Theorem~\ref{thm:chains-to-trees} for the special case
\(N=2^D\) by induction on the hierarchy of depth \(D\). Throughout the proof,
we identify the boundary variables as \(\bz_0=\bx\) and \(\bz_N=\by\).

For a chain interval \([l,r]\), let \(n\) denote its midpoint. By Bayes'
rule and the first-order Markov property, the posterior $p_\theta(\bz_n\mid\bz_l,\bz_r)$, for any $l<n<r$ is
\begin{equation}
\beta_{l,n,r} := p_\theta(\bz_n\mid\bz_l,\bz_r)
=
\frac{
p_\theta(\bz_n\mid\bz_l)\,
p_\theta(\bz_r\mid\bz_n)
}{
p_\theta(\bz_r\mid\bz_l)
} \;.
\label{eqn:local-bridge-posterior}
\end{equation}
We show that for any chain of length $N$, the posterior $p_\theta(\bz_{1:N-1}\mid\bz_0,\bz_N)$ can be constructed recursively, to give a product of $N-1$ terms of the form Eq.~\eqref{eqn:local-bridge-posterior}.

\paragraph{Base case:} We denote by \(\mathcal{N}_D\) the tree factorization of a chain interval
with \(2^D+1\) boundary and intermediate variables. The smallest nontrivial
case, \(\mathcal{N}_1\), contains 3 variables
\(\bz_0,\bz_1,\bz_2\), corresponding to a network with one hidden block  $\bz_0 \rightarrow \bz_1 \rightarrow \bz_2$ ($\bx \rightarrow \bz_1 \rightarrow \by$). The posterior :
\begin{equation*}
\mathcal{N}_1:\quad
p_\theta(\bz_1\mid\bz_0,\bz_2)
=
\frac{
p_\theta(\bz_1\mid\bz_0)\,
p_\theta(\bz_2\mid\bz_1)
}{
p_\theta(\bz_2\mid\bz_0)
} = \beta_{0,1,2}\,.
\end{equation*}
Thus, for \(D=1\), the chain posterior and the tree factorization coincide.

The next case, \(\mathcal{N}_2\), i.e. a network \(\bz_0 \rightarrow \bz_1 \rightarrow \dots \rightarrow \bz_4\), can be constructed recursively form 2 networks of the form \(\mathcal{N}_1\). The midpoint \(\bz_2\) splits the
chain into the subintervals \([0,2]\) and \([2,4]\). Therefore,
\begin{equation*}
\begin{aligned}
&\mathcal{N}_2:\quad p_\theta(\bz_{1:3}\mid\bz_0,\bz_4) \, = \frac{p_\theta(\bz_{1:4}\mid\bz_0)}{p_\theta(\bz_4\mid\bz_0)} \\
& \quad =
\frac{p_\theta(\bz_1\mid\bz_0)\,p_\theta(\bz_2\mid\bz_1)\,p_\theta(\bz_3\mid\bz_2)\,p_\theta(\bz_4\mid\bz_3)}{p_\theta(\bz_4\mid\bz_0)} \\
& \quad =
\frac{p_\theta(\bz_1, \bz_2 \mid\bz_0)\, p_\theta(\bz_3, \bz_4 \mid\bz_2)}{p_\theta(\bz_4\mid\bz_0)} \\
& \quad =
\frac{\color{red} p_\theta(\bz_2 \mid\bz_0)\,p_\theta(\bz_4 \mid\bz_2)}{p_\theta(\bz_4\mid\bz_0)}\,
\frac{p_\theta(\bz_1, \bz_2 \mid\bz_0)}{\color{blue} p_\theta(\bz_2 \mid\bz_0)}\,
\frac{p_\theta(\bz_3, \bz_4 \mid\bz_2)}{\color{blue} p_\theta(\bz_4 \mid\bz_2)} \\
& \quad =
p_\theta(\bz_2\mid\bz_0,\bz_4) \times p_\theta(\bz_1\mid\bz_0,\bz_2)
\times
p_\theta(\bz_3\mid\bz_2,\bz_4)\\
& \quad =\,\beta_{0,2,4}\times\beta_{0,1,2}\times\beta_{2,3,4}\,,
\end{aligned}
\end{equation*}
where red and blue terms cancel to 1.
This is the posterior at the \(\mathcal{N}_2\)-root ($\bz_2$) followed by two \(\mathcal{N}_1\)-typed networks
 for the left and right subintervals.

Similarly, \(\mathcal{N}_3\) contains variables
\(\bz_0,\bz_1,\dots,\bz_8\). Its root midpoint is \(\bz_4\), yielding
\begin{equation}
\begin{aligned}
\mathcal{N}_3:\quad
p_\theta(\bz_{1:7}\mid\bz_0,\bz_8)
&=
\overbrace{p_\theta(\bz_4\mid\bz_0,\bz_8)}^{\beta_{0,4,8}}
\\
&\quad\times
p_\theta(\bz_{1:3}\mid\bz_0,\bz_4)
\\
&\quad\times
p_\theta(\bz_{5:7}\mid\bz_4,\bz_8).
\end{aligned}
\label{eqn:N3}
\end{equation}
Substituting \(\mathcal{N}_2\)-type factorizations for the left and right 
subintervals
\begin{equation*}
\begin{aligned}
\mathcal{N}_3:\quad
p_\theta(\bz_{1:7}\mid\bz_0,\bz_8)
=
\beta_{0,4,8}\times
\beta_{0,2,4}\times
\beta_{4,6,8}\times\\
\beta_{0,1,2}\times
\beta_{2,3,4}\times
\beta_{4,5,6}\times
\beta_{6,7,8}
\end{aligned}
\end{equation*}
shows that \(\mathcal{N}_3\) is again a product of of tree-structured posterior factors.

\paragraph{Induction step:} Now we show the general case $\mathcal{N}_D$. Assuming that the result above holds for every chain interval of hierarchy depth
\(D-1\), we show the iteration $\mathcal{N}_{D-1} \rightarrow \mathcal{N}_D$, for all $D>2$. Consider the interval \([0,N]\), where \(N=2^D\), and let \(m=N/2\). Conditioning on the midpoint \(\bz_m\) separates the left and
right subchains by the first-order Markov property, analogous to the $\mathcal{N}_2$ case demonstrated above:
\begin{equation*}
\begin{aligned}
&\mathcal{N}_D:\quad
p_\theta(\bz_{1:N-1}\mid\bz_0,\bz_N)
=  \frac{p_\theta(\bz_{1:N}\mid\bz_0)}{p_\theta(\bz_N\mid\bz_0)} \\
& =
\frac{p_\theta(\bz_{1:m} \mid\bz_0)\, p_\theta(\bz_{m+1:N} \mid\bz_m)}{p_\theta(\bz_N\mid\bz_0)} \\
& =
\frac{\color{red} p_\theta(\bz_m \mid\bz_0)\,p_\theta(\bz_N \mid\bz_m)}{p_\theta(\bz_N\mid\bz_0)}\,
\frac{p_\theta(\bz_{1:m} \mid\bz_0)}{\color{blue} p_\theta(\bz_m \mid\bz_0)}\,
\frac{p_\theta(\bz_{m+1:N} \mid\bz_m)}{\color{blue} p_\theta(\bz_N \mid\bz_m)} \\
& =
p_\theta(\bz_m\mid\bz_0,\bz_N)\times\\
&\qquad p_\theta(\bz_{1:m-1}\mid\bz_0,\bz_m)\times
 p_\theta(\bz_{m+1:N-1}\mid\bz_m,\bz_N)\\
 & =\,\beta_{0,m,N}\times\\
&\qquad p_\theta(\bz_{1:m-1}\mid\bz_0,\bz_m)\times
 p_\theta(\bz_{m+1:N-1}\mid\bz_m,\bz_N)\;.
\end{aligned}
\label{eqn:ND}
\end{equation*}
The first term $\beta_{0,m,N}$ is the posterior at the root. The second and third
terms are posteriors over subintervals of hierarchy depth \(D-1\). By the
induction hypothesis, each decomposes into tree-structured
posterior factors. 
Hence, the depth-\(D\) posterior also factorizes to tree-structured
posterior factors.

Therefore, by induction,
\begin{equation}
\begin{aligned}
p_\theta(\bz_{1:N-1}\mid\bz_0,\bz_N)
&=
\prod_{n=1}^{N-1}
\beta_{\ell(n),n,r(n)} \\
&=
\prod_{n=1}^{N-1}
p_\theta\!\left(
\bz_n\mid\operatorname{pa}(n)
\right).
\label{eqn:appendix-chains-to-trees}
\end{aligned}
\end{equation}
Here, \(\ell(n)\) and \(r(n)\) denote the left and right boundary indices
associated with node \(n\), and
\(\operatorname{pa}(n)
=
\bigl(\bz_{\ell(n)},\bz_{r(n)}\bigr)\)
denotes the corresponding boundary representations.

Identifying \(\bz_0=\bx\) and \(\bz_N=\by\) gives the factorization stated
in Theorem~\ref{thm:chains-to-trees}.

\paragraph{Arbitrary chain lengths.}
For arbitrary \(N\), the same argument applies using unequal recursive
splits. Given an interval \([l,r]\) with \(r-l\geq 2\), choose
\[
n
=
\left\lfloor
\frac{l+r}{2}
\right\rfloor .
\]
Since \(l<n<r\), the variables
\(\bz_l\rightarrow\bz_n\rightarrow\bz_r\) form a valid subchain, and the
posterior factorization in Eq.~\eqref{eqn:local-bridge-posterior} remains valid.
Recursively applying the construction to the resulting subintervals yields
the corresponding tree-structured posterior factorization with hierarchy
depth at most \(\lceil\log_2 N\rceil\).
\qed

It is worth noting, that the construction presented above is not an approximation but a different tree-structured representation of the true posterior $p_\theta(\bz_{1:N-1}\mid\bz_0,\bz_N)$. The intractable normalizers (e.g. $p_\theta(\by\mid\bx)$) are still present in this construction but distributed over the entire tree. A variational approximation that carefully follows the same tree-structured construction, can closely match the true posterior by learning local approximations to intractable normalizers as shown in theory and empirically in the following sections and in the main text.

\subsection{Derivation of the Variational Objective}
\label{app:proof-variational-objective}

\begin{theorem}[Variational tree learning objective]
\label{thm:variational-tree-objective}
Let \(p_\theta(\bz_{1:N-1},\by\mid\bx)\) be the Markov-chain model in
Eq.~\eqref{eqn:conditional-chain}, and let
\(q_\phi(\bz_{1:N-1}\mid\bx,\by)\) be the variational posterior
in Eq.~\eqref{eqn:tree-variational-posterior}. 
Then the negative log-likelihood admits the variational upper bound
\begin{equation}
\mathcal{L}(\theta)
=
-\log p_\theta(\by\mid\bx)
\leq
\mathcal{F}(\theta,\phi),
\label{eqn:variational-bound}
\end{equation}
where
\begin{equation}
\begin{aligned}
\mathcal{F}(\theta,\phi)
&=
\mathbb{E}_{q_\phi}\!
\Bigg[
\sum_{n=1}^{N-1}
\log
\frac{
q_{\phi_n}\!\left(\bz_n\mid\operatorname{pa}(n)\right)
}{
p_{\theta_n}(\bz_n\mid\bz_{n-1})
}
\\
&\qquad\qquad
-
\log p_{\theta_N}(\by\mid\bz_{N-1})
\Bigg].
\end{aligned}
\end{equation}
\end{theorem}

We derive the variational upper bound stated in
Theorem~\ref{thm:variational-tree-objective}. For compactness, let
\[
q_\phi
=
q_\phi(\bz_{1:N-1}\mid\bx,\by),
\qquad
p_\theta^{\mathrm{post}}
=
p_\theta(\bz_{1:N-1}\mid\bx,\by).
\]

We first make explicit the tree-structured factorization of the variational
posterior. In the special case \(N=2^D\), the root representation is
\(\bz_{N/2}\), conditioned on the boundary variables \(\bx\) and \(\by\).
The root splits the chain into the intervals \([0,N/2]\) and \([N/2,N]\),
whose midpoint representations are \(\bz_{N/4}\) and \(\bz_{3N/4}\),
respectively. Hence, the first levels of the factorization are
\begin{equation}
\begin{aligned}
q_\phi(\bz_{1:N-1}\mid\bx,\by)
&=
q_{\phi_{N/2}}
\!\left(
\bz_{N/2}\mid\bx,\by
\right)
\\
&\quad\times
q_{\phi_{N/4}}
\!\left(
\bz_{N/4}\mid\bx,\bz_{N/2}
\right)
\\
&\quad\times
q_{\phi_{3N/4}}
\!\left(
\bz_{3N/4}\mid\bz_{N/2},\by
\right)
\\
&\quad\times\cdots .
\end{aligned}
\label{eqn:appendix-explicit-q-factorization}
\end{equation}
Recursively applying the same construction to all remaining subintervals
gives the compact product form
\begin{equation}
q_\phi(\bz_{1:N-1}\mid\bx,\by)
=
\prod_{n=1}^{N-1}
q_{\phi_n}\!\left(
\bz_n\mid\operatorname{pa}(n)
\right).
\label{eqn:appendix-q-factorization}
\end{equation}
Here \(\operatorname{pa}(n)=(\bz_l,\bz_r)\) denotes the two boundary representations associated with node \(n\). 
where \(\bz_l\) and \(\bz_r\) are the boundary representations of the
current interval.
This construction yields a hierarchy of depth at most
\(\lceil\log_2 N\rceil\).

For a single training example, the negative log-likelihood is
\begin{equation}
\mathcal{L}(\theta)
=
-\log p_\theta(\by\mid\bx).
\label{eqn:appendix-negative-log-likelihood}
\end{equation}
By Jensen's inequality we obtain 
\begin{equation}
\begin{aligned}
\mathcal{L}(\theta)
&\leq
-\log p_\theta(\by\mid\bx)
+
D_{\mathrm{KL}}\!\left(
q_\phi
\,\middle\|\,
p_\theta^{\mathrm{post}}
\right)
\\
&=
-\log p_\theta(\by\mid\bx)
\\
&\quad+
\mathbb{E}_{q_\phi}
\left[
\log
\frac{
q_\phi(\bz_{1:N-1}\mid\bx,\by)
}{
p_\theta(\bz_{1:N-1}\mid\bx,\by)
}
\right]
\\
&=
\mathbb{E}_{q_\phi}
\left[
\log
\frac{
q_\phi(\bz_{1:N-1}\mid\bx,\by)
}{
p_\theta(\bz_{1:N-1},\by\mid\bx)
}
\right]
\\
&=:
\mathcal{F}(\theta,\phi).
\end{aligned}
\label{eqn:appendix-variational-bound}
\end{equation}
The third equality uses Eq.~\eqref{eqn:true-posterior}, i.e.,
\[
p_\theta(\bz_{1:N-1}\mid\bx,\by)
=
\frac{
p_\theta(\bz_{1:N-1},\by\mid\bx)
}{
p_\theta(\by\mid\bx)
}.
\]
Using the Markov-chain factorization
\begin{equation}
\begin{aligned}
p_\theta(\bz_{1:N-1},\by\mid\bx)
&=
p_{\theta_N}(\by\mid\bz_{N-1})
\\
&\quad\times
\prod_{n=1}^{N-1}
p_{\theta_n}(\bz_n\mid\bz_{n-1}),
\end{aligned}
\label{eqn:appendix-markov-factorization}
\end{equation}
together with the tree-structured variational factorization in
Eq.~\eqref{eqn:appendix-q-factorization}, we obtain
\begin{equation}
\begin{aligned}
\mathcal{F}(\theta,\phi)
&=
\mathbb{E}_{q_\phi}
\Bigg[
\log
\frac{
\prod_{n=1}^{N-1}
q_{\phi_n}\!\left(
\bz_n\mid\operatorname{pa}(n)
\right)
}{
p_{\theta_N}(\by\mid\bz_{N-1})
\prod_{n=1}^{N-1}
p_{\theta_n}(\bz_n\mid\bz_{n-1})
}
\Bigg]
\\
&=
\mathbb{E}_{q_\phi}
\Bigg[
\sum_{n=1}^{N-1}
\log
\frac{
q_{\phi_n}\!\left(
\bz_n\mid\operatorname{pa}(n)
\right)
}{
p_{\theta_n}(\bz_n\mid\bz_{n-1})
}
\\
&\qquad\qquad
-
\log p_{\theta_N}(\by\mid\bz_{N-1})
\Bigg].
\end{aligned}
\label{eqn:appendix-tree-objective}
\end{equation}

Thus, the negative log-likelihood is upper-bounded by a variational objective that decomposes into node-wise local terms associated with the tree-structured variational posterior, together with the output likelihood term. 
This is precisely the tree-structured variational objective stated in
Theorem~\ref{thm:variational-tree-objective}.
\qed

\subsection{Node-Wise Loss Estimation}
\label{app:nodewise-losses}

For a sample from the tree-structured variational posterior, the
node-wise contribution associated with \(\bz_n\) is
\begin{equation}
\ell_n
=
\log
\frac{
q_{\phi_n}\!\left(
\bz_n\mid\operatorname{pa}(n)
\right)
}{
p_{\theta_n}\!\left(
\bz_n\mid\bz_{n-1}
\right)
}.
\label{eqn:appendix-nodewise-loss}
\end{equation}
For fixed conditioning representations, the expectation of this
log-ratio under the tree-structured variational factor is
\begin{equation}
\mathbb{E}_{
q_{\phi_n}(\bz_n\mid\operatorname{pa}(n))
}
\left[
\ell_n
\right]
=
D_{\mathrm{KL}}
\left(
q_{\phi_n}\!\left(
\bz_n\mid\operatorname{pa}(n)
\right)
\,\middle\|\,
p_{\theta_n}\!\left(
\bz_n\mid\bz_{n-1}
\right)
\right).
\label{eqn:appendix-nodewise-kl}
\end{equation}

For the MNIST experiments, the latent representations are modeled using
factorized Bernoulli distributions:
\begin{align}
q_{\phi_n}\!\left(
\bz_n\mid\operatorname{pa}(n)
\right)
&=
\operatorname{Bernoulli}
\left(
\boldsymbol{\pi}_n
\right),
\\
p_{\theta_n}\!\left(
\bz_n\mid\bz_{n-1}
\right)
&=
\operatorname{Bernoulli}
\left(
\boldsymbol{\rho}_n
\right),
\end{align}
where \(\boldsymbol{\pi}_n\) and \(\boldsymbol{\rho}_n\) denote the
inferred and forward-predicted Bernoulli probabilities, respectively.
The corresponding node-wise KL contribution is
\begin{equation}
\begin{aligned}
D_{\mathrm{KL}}
\left(
q_{\phi_n}\,\middle\|\,p_{\theta_n}
\right)
=
\sum_j
\Bigg[
&\pi_{n,j}
\log
\frac{\pi_{n,j}}{\rho_{n,j}}
\\
&+
\left(1-\pi_{n,j}\right)
\log
\frac{1-\pi_{n,j}}{1-\rho_{n,j}}
\Bigg].
\end{aligned}
\label{eqn:appendix-bernoulli-kl}
\end{equation}

For the vision and language-modeling experiments, the corresponding
distributions are modeled using fixed-variance isotropic Gaussians. The
tree-structured variational factor is parameterized by the inferred mean
\(\boldsymbol{\mu}_n\), whereas the forward transition is parameterized
by the prediction \(\boldsymbol{\mu}^{\rightarrow}_n\):
\begin{align}
q_{\phi_n}\!\left(
\bz_n\mid\operatorname{pa}(n)
\right)
&=
\mathcal{N}\!\left(
\bz_n
\,\middle|\,
\boldsymbol{\mu}_n,
\frac{\sigma^2}{2}\mathbf{I}
\right),
\\
p_{\theta_n}\!\left(
\bz_n\mid\bz_{n-1}
\right)
&=
\mathcal{N}\!\left(
\bz_n
\,\middle|\,
\boldsymbol{\mu}^{\rightarrow}_n,
\sigma^2\mathbf{I}
\right).
\end{align}
The Gaussian KL divergence is
\begin{equation}
\begin{aligned}
D_{\mathrm{KL}}
\left(
q_{\phi_n}\,\middle\|\,p_{\theta_n}
\right)
=
\frac{1}{2}
\Bigg[
&
d
\left(
\frac{1}{2}
-1
+
\log 2
\right)
\\
&+
\frac{
\left\|
\boldsymbol{\mu}_n
-
\boldsymbol{\mu}^{\rightarrow}_n
\right\|_2^2
}{
\sigma^2
}
\Bigg],
\end{aligned}
\label{eqn:appendix-gaussian-kl}
\end{equation}
where \(d=\dim(\bz_n)\).

For the distributions used in our experiments, these node-wise
divergences reduce to standard training losses.
For fixed-variance Gaussian representations, the mean-dependent term in Eq.~\eqref{eqn:appendix-gaussian-kl} is a variance-scaled squared-error loss between the inferred and forward-predicted means.
For Bernoulli representations, Eq.~\eqref{eqn:appendix-bernoulli-kl} is implemented using binary cross-entropy between the inferred and forward-predicted Bernoulli probabilities.
For nodes whose right parent is the output node \(\by\), the right-boundary block is a classifier rather than a latent-state transition, and the corresponding node-wise loss is the standard multiclass cross-entropy loss.

The expectation in the complete TreeProp objective is estimated using
samples from the tree-structured variational posterior:
\begin{equation}
\bz_{1:N-1}^{(k)}
\sim
q_\phi
\left(
\bz_{1:N-1}\mid\bx,\by
\right),
\qquad
k=1,\ldots,K.
\label{eqn:appendix-mc-samples}
\end{equation}
This gives the Monte Carlo estimator
\begin{equation}
\mathcal{F}(\theta,\phi)
\approx
\frac{1}{K}
\sum_{k=1}^{K}
\left[
\sum_{n=1}^{N-1}
\ell_n^{(k)}
-
\log
p_{\theta_N}
\left(
\by\mid\bz_{N-1}^{(k)}
\right)
\right].
\label{eqn:appendix-mc-objective}
\end{equation}
We use one Monte Carlo sample, \(K=1\), for each training example in all
reported experiments. This provides a stochastic estimate of the
variational objective while avoiding the additional computation and
memory required for multiple posterior samples.

\subsection{Gradient Bias and Descent of the TreeProp Objective}
\label{app:TreeProp-gradient-bias}

The preceding subsection defined the node-wise losses that TreeProp actually
evaluates. The TreeProp update is obtained by
differentiating the tree objective through the tree computation graph, with the
right-boundary correction \(\boldsymbol\mu_n^{\leftarrow}\) of
Eq.~\eqref{eqn:posterior-mean} treated as a constant. Two questions follow. First, how far does the resulting update sit
from the true gradient? We show that the difference decomposes into a score term
and a path term, both controlled by divergences
\(D_{\mathrm{KL}}(q_{n}\|p_{n})\)
between the variational and the exact
posterior on each chain block, defined below and driven to zero by the
node-wise local losses. Second, when does following it still decrease \(\mathcal L\)?
We give a sufficient condition in terms of the same divergences. The subsection
closes with a direct empirical test of both statements along a training run.

Consider a fixed training pair \((\bx,\by)\), and let
\(\bz=\bz_{1:N-1}\) collect all intermediate latent variables. Define
\[
\mathcal L(\theta)
=
-\log p_\theta(\by\mid\bx)
\]
and
\begin{align}
\mathcal F(\theta,\phi)
&=
\mathcal L(\theta)
+
D_{\mathrm{KL}}
\left(
q_{\phi}(\bz\mid\bx,\by)
\middle\|
p_\theta(\bz\mid\bx,\by)
\right)\nonumber \\
&=
\mathbb E_{q_{\phi}(\bz\mid\bx,\by)}
\left[
\log
\frac{
q_{\phi}(\bz\mid\bx,\by)
}{
p_\theta(\bz,\by\mid\bx)
}
\right].
\label{eqn:appendix-TreeProp-objective}
\end{align}
Here \(q_\phi\) is the tree-structured variational posterior of
Eq.~\eqref{eqn:tree-variational-posterior}, whose factors are constructed as in
Eq.~\eqref{eqn:posterior-mean}. Recall that \(\phi\) denotes the full set of
variational parameters and contains the network
parameters \(\theta\) together with the auxiliary tree-block parameters. We
retain the notation \(\mathcal F(\theta,\phi)\) of
Eq.~\eqref{eqn:tree-variational-objective}, in which the first argument records
that the model term \(\mathcal L(\theta)\) depends on \(\theta\) alone while the
variational term depends on all of \(\phi\). This
\(\theta\)-dependence of \(q_\phi\) is precisely what the term \(S_q\) below
captures, and it is the source of the second bias component in
Eq.~\eqref{eqn:appendix-total-bias-decomposition}.

Throughout, \(\widetilde{\nabla}_\theta\) denotes differentiation through the
training computation graph, treating stop-gradient quantities as constants.
On explicit forward-model terms, it coincides with the ordinary derivative
\(\nabla_\theta\). Thus, the derivative of \(q_{\phi}\) includes its
differentiable paths but not its stopped inversion-feedback path.

Assume that \(p_\theta\) and \(q_{\phi}\) have differentiable densities
with common, parameter-independent support, that differentiation and
integration may be interchanged, and that the required first and second
moments are finite.

Write
\[
p(\bz)
=
p_\theta(\bz\mid\bx,\by),
\quad
q(\bz)
=
q_{\phi}(\bz\mid\bx,\by),
\]
and define
\[
s_\theta(\bz)
=
\nabla_\theta
\log p_\theta(\bz,\by\mid\bx),
\;
S_q(\bz)
=
\widetilde{\nabla}_\theta
\log q_{\phi}(\bz\mid\bx,\by),
\]
together with
\[
r(\bz)
=
\log\frac{q(\bz)}{p(\bz)}.
\]
Both \(p\) and \(q\) are densities over the \emph{joint} latent vector
\(\bz=\bz_{1:N-1}\), not over an individual node; \(p\) is the exact posterior of
the Markov-chain model and \(q\) the tree-structured variational posterior of
Eq.~\eqref{eqn:tree-variational-posterior}. 

\paragraph{Exact gradient decomposition.}

By the Fisher identity,
\begin{equation}
g
:=
\nabla_\theta\mathcal L(\theta)
=
-\mathbb E_p[s_\theta(\bz)].
\label{eqn:appendix-true-gradient}
\end{equation}
Differentiating Eq.~\eqref{eqn:appendix-TreeProp-objective} through the active
computation graph gives
\[
\begin{aligned}
\widetilde{\nabla}_\theta\mathcal F
&=
\int
\widetilde{\nabla}_\theta q(\bz)
\left[
\log q(\bz)
-
\log p_\theta(\bz,\by\mid\bx)
\right]
d\bz
-
\mathbb E_q[s_\theta],
\end{aligned}
\]
because
\[
\mathbb E_q[S_q]
=
\int
\widetilde{\nabla}_\theta q(\bz)\,d\bz
=
\widetilde{\nabla}_\theta
\int q(\bz)\,d\bz
=
0.
\]
Using
\[
\log p_\theta(\bz,\by\mid\bx)
=
\log p_\theta(\bz\mid\bx,\by)
+
\log p_\theta(\by\mid\bx),
\]
the evidence term vanishes after integration against
\(\widetilde{\nabla}_\theta q\). Therefore,
\[
\int
\widetilde{\nabla}_\theta q(\bz)r(\bz)\,d\bz
=
\mathbb E_q[S_qr]
=
\operatorname{Cov}_q(S_q,r),
\]
and hence
\begin{equation}
\widehat g
:=
\widetilde{\nabla}_\theta\mathcal F(\theta,\phi)
=
-\mathbb E_q[s_\theta]
+
\operatorname{Cov}_q(S_q,r).
\label{eqn:appendix-total-gradient}
\end{equation}

Thus, the total gradient bias \(b=\widehat g-g\) decomposes as
\begin{equation}
b
=
\underbrace{
\mathbb E_p[s_\theta]
-
\mathbb E_q[s_\theta]
}_{b_{\mathrm{score}}}
+
\underbrace{
\operatorname{Cov}_q(S_q,r)
}_{b_{\mathrm{path}}}.
\label{eqn:appendix-total-bias-decomposition}
\end{equation}
The first term is the posterior score mismatch; the second is the contribution
from the active \(\theta\)-dependence of the variational construction. If \(q\)
is held fixed under \(\widetilde{\nabla}_\theta\), then \(S_q=0\), recovering
the original fixed-\(q\) result.

If \(q=p\), then both the score mismatch and \(r\) vanish, so
\begin{equation}
q=p
\quad\Longrightarrow\quad
b=0.
\label{eqn:appendix-exact-posterior-unbiased}
\end{equation}
This is consistent with Theorem~\ref{thm:chains-to-trees}, which establishes
that the exact posterior is realizable within the tree family, and with its
numerical verification. Stationarity of \(\mathcal F\) in the auxiliary directions alone
is not sufficient: it does not by itself imply \(b_{\mathrm{path}}=0\), because
\(q_\phi\) retains an active \(\theta\)-dependence. More precisely, the
covariance term vanishes whenever
\[
\mathbb E_q
\left[
S_q(\bz)r(\bz)
\right]
=
0,
\]
that is, whenever the objective is stationary along the distributional
direction induced by the active \(\theta\)-dependence of \(q\). Exact posterior
matching, \(q=p\), is a sufficient condition; approximate matching gives the
quantitative bounds below.

\paragraph{Bias bounds.}

Assume first that
\[
\|s_\theta(\bz)\|
\leq
S
\qquad
\text{for all }\bz.
\]
This is a bounded-score assumption for some finite real number $S$. While $S$ may in theory be unbounded, empirically we find it takes a small finite value. Then
\[
\begin{aligned}
\|b_{\mathrm{score}}\|
&\leq
S
\int
|p(\bz)-q(\bz)|
\,d\bz
\\
&=
2S\,\operatorname{TV}(q,p)
\\
&\leq
S
\sqrt{
2D_{\mathrm{KL}}(q\|p)
},
\end{aligned}
\]
where the last step follows from Pinsker's inequality, and
\(\operatorname{TV}(q,p)\) denotes the total variation distance between
\(q\) and \(p\).

The bound above is stated in the joint divergence \(D_{\mathrm{KL}}(q\|p)\) over
all of \(\bz_{1:N-1}\), which is neither tractable nor the quantity that any
local loss minimizes. The following refinement descends to per-block
divergences by exploiting the fact that each block's score depends on only two
adjacent latents.

For a sharper local result, suppose
\(\theta=(\theta_1,\ldots,\theta_N)\) consists of disjoint parameter blocks, one
per chain transition. Throughout, \emph{block} refers to these \(N\) transitions
and \emph{node} to the latents \(\bz_n\) they connect; the auxiliary tree blocks
of \(\phi\) are not indexed here. The index \(n\in\{1,\dots,N\}\) runs over
blocks, with \(\theta_n\) parameterizing the transition into \(\bz_n\) and block
\(N\) the output map into \(\bz_N=\by\); \(\mathbf{u}_n\) below collects the
latents that \(\theta_n\) touches.
For a hidden transition block, define
\[
s_n(\bz)
=
\nabla_{\theta_n}
\log p_{\theta_n}(\bz_n\mid\bz_{n-1}),
\qquad
\mathbf{u}_n
=
(\bz_{n-1},\bz_n),
\]
while for the output block,
\[
s_N(\bz)
=
\nabla_{\theta_N}
\log p_{\theta_N}(\by\mid\bz_{N-1}),
\qquad
\mathbf{u}_N
=
\bz_{N-1}.
\]
Assume
\[
\|s_n(\mathbf{u}_n)\|
\leq
S_n.
\]
Let \(q_n\) and \(p_n\) denote the marginals of \(q\) and \(p\) on
\(\mathbf{u}_n\), that is, on \((\bz_{n-1},\bz_n)\) for a hidden block and on
\(\bz_{N-1}\) for the output block. Since \(s_n\) is a function of
\(\mathbf{u}_n\) alone, only these marginals enter \(b_n^{\mathrm{score}}\),
which is what makes the refinement below possible. Define
\[
b_n^{\mathrm{score}}
=
\mathbb E_{p_n}[s_n]
-
\mathbb E_{q_n}[s_n].
\]
Then
\begin{equation}
\|b_n^{\mathrm{score}}\|
\leq
S_n
\sqrt{
2D_{\mathrm{KL}}
\left(
q_n
\middle\|
p_n
\right)
}.
\label{eqn:appendix-local-score-bound}
\end{equation}

\begin{figure}[t]
\centering
\includegraphics[width=\columnwidth]{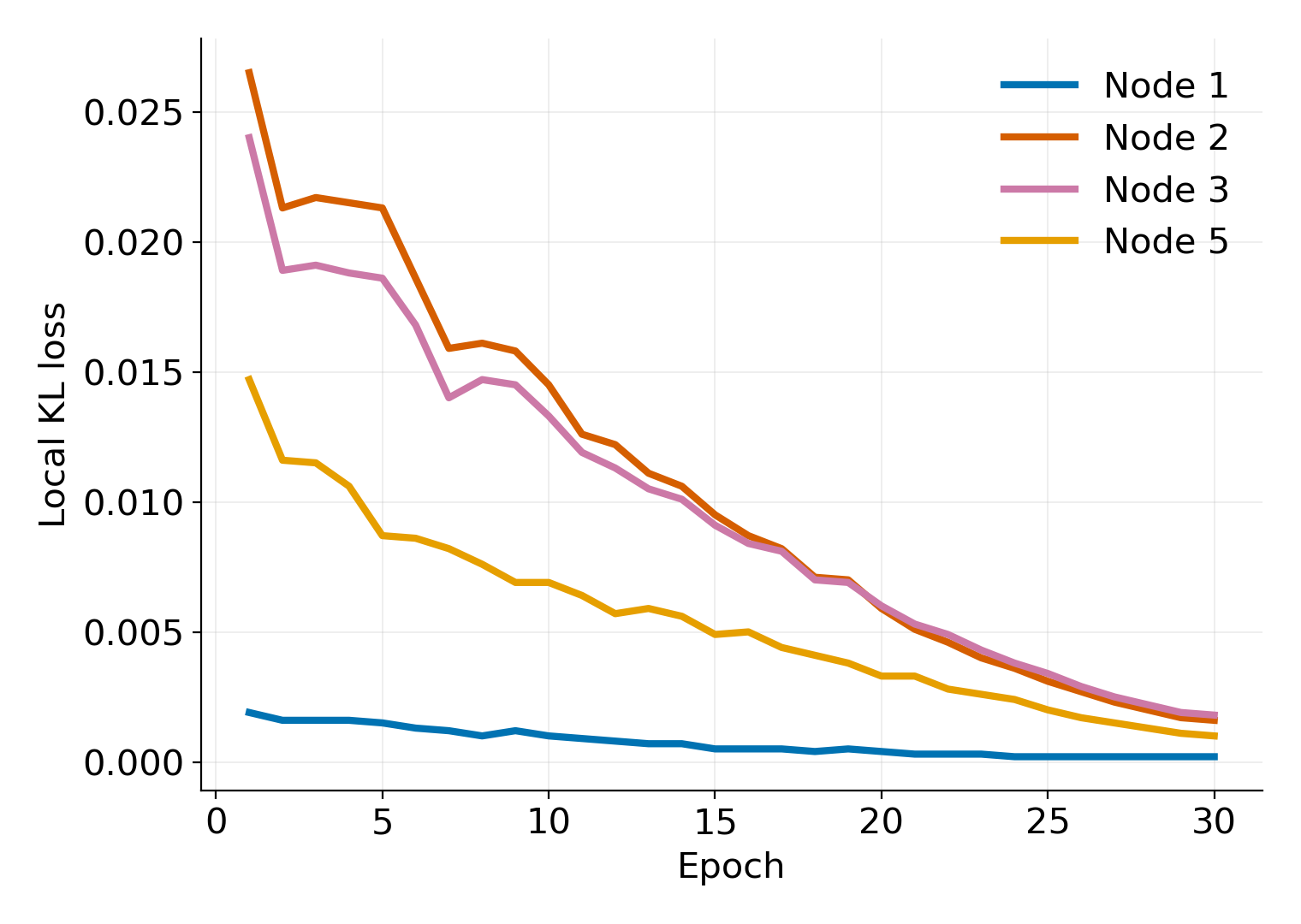}
\caption{Per-node local KL consistency losses over training (depth-3,
WikiText-103). These terms control the bias of the TreeProp update; all decay to near zero.}
\label{fig:local_kl}
\end{figure}

These are not the node-wise factors of
Appendix~\ref{app:nodewise-losses}: \(q_{\phi_n}\) and \(p_{\theta_n}\) there are
conditionals, the variational factor at node \(n\) and the model transition,
whereas \(q_n\) and \(p_n\) here are marginals of the joint posteriors
\[
p_n := p_\theta(\bz_{n-1}, \bz_n\mid\bx,\by) =  \sum_{\bz \setminus (\bz_{n-1}, \bz_n)} p_\theta(\bz\mid\bx,\by)\;,
\]
\[
q_n := q_\phi(\bz_{n-1}, \bz_n\mid\bx,\by) = \sum_{\bz \setminus (\bz_{n-1}, \bz_n)} q_\phi(\bz\mid\bx,\by)\;.
\]
The two are nonetheless linked, through marginalization over $\bz_{n-1}$. Driving the local terms down minimises
\(\mathcal F\), and since
\(\mathcal F=\mathcal L(\theta)+D_{\mathrm{KL}}(q\|p)\) with
\(\mathcal L(\theta)\) independent of the variational parameters, it drives
\(D_{\mathrm{KL}}(q\|p)\to 0\); marginalisation cannot increase divergence, so
each \(D_{\mathrm{KL}}(q_n\|p_n)\to 0\) in turn. This is what allows reading Figure~\ref{fig:local_kl} as evidence about the bias.

Since the parameter blocks are disjoint,
\[
\|b_{\mathrm{score}}\|^2
=
\sum_{n=1}^{N}
\|b_n^{\mathrm{score}}\|^2,
\]
and therefore
\begin{equation}
\|b_{\mathrm{score}}\|
\leq
\left[
2
\sum_{n=1}^{N}
S_n^2
D_{\mathrm{KL}}
\left(
q_n
\middle\|
p_n
\right)
\right]^{1/2}.
\label{eqn:appendix-local-global-score-bound}
\end{equation}

Since marginalization cannot increase KL divergence,
\[
D_{\mathrm{KL}}
\left(
q_n
\middle\|
p_n
\right)
\leq
D_{\mathrm{KL}}(q\|p).
\]
Thus, defining
\[
S_{\mathrm{glob}}
=
\left(
\sum_{n=1}^{N}
S_n^2
\right)^{1/2},
\]
we obtain
\begin{equation}
\|b_{\mathrm{score}}\|
\leq
S_{\mathrm{glob}}
\sqrt{
2D_{\mathrm{KL}}(q\|p)
}.
\label{eqn:appendix-global-score-bound}
\end{equation}
If the per-block bounds admit a uniform bound \(S_0\), that is
\(S_n\leq S_0\) for every \(n\), then
\[
S_{\mathrm{glob}}
\leq
\sqrt N\,S_0.
\]

The refinement applies to \(b_{\mathrm{score}}\) but not to
\(b_{\mathrm{path}}\), and the asymmetry is structural. The score term
decomposes because \(\theta\) decomposes: each \(s_n\) depends only on
\(\mathbf{u}_n\), so marginalization bounds it by divergences that the local
losses directly minimize, namely those measured in
Figure~\ref{fig:local_kl}. No corresponding decomposition is available for
\(b_{\mathrm{path}}\), since \(r=\log(q/p)\) is the full joint log-density ratio
and the variational parameters \(\phi\) are not partitioned along the chain. The
auxiliary parameters therefore control the magnitude of both terms, through
\(q_n\), \(\operatorname{Var}_q(r)\) and \(G_q\) (both defined below), but only the
\(\theta\)-side inherits the chain's block structure.

Define
\[
G_q^2
=
\mathbb E_q
\|S_q(\bz)\|^2.
\]
Since \(\mathbb E_q[S_q]=0\), Cauchy--Schwarz gives
\begin{equation}
\|b_{\mathrm{path}}\|
\leq
G_q
\sqrt{
\operatorname{Var}_q(r)
}.
\label{eqn:appendix-path-bound}
\end{equation}
Combining the two terms,
\begin{equation}
\|b\|
\leq
S
\sqrt{
2D_{\mathrm{KL}}(q\|p)
}
+
G_q
\sqrt{
\operatorname{Var}_q(r)
},
\label{eqn:appendix-total-bias-bound}
\end{equation}
or, using the local marginal divergences,
\begin{equation}
\begin{aligned}
\|b\|
&\leq
\left[
2
\sum_{n=1}^{N}
S_n^2
D_{\mathrm{KL}}
\left(
q_n
\middle\|
p_n
\right)
\right]^{1/2}
\\
&\quad+
G_q
\sqrt{
\operatorname{Var}_q(r)
}.
\end{aligned}
\label{eqn:appendix-total-bias-local}
\end{equation}

This triangle-inequality bound is conservative. Indeed,
\[
\|b\|^2
=
\|b_{\mathrm{score}}\|^2
+
\|b_{\mathrm{path}}\|^2
+
2
\langle
b_{\mathrm{score}},
b_{\mathrm{path}}
\rangle,
\]
so the two components may partially cancel, as observed empirically. We note this bound is loose, but does prove that as $q$ and $p$ get closer, the bias must also converge to zero.
\paragraph{Equal-covariance Gaussian specialization.}

If
\[
q
=
\mathcal N(\boldsymbol\mu_q,\Sigma),
\qquad
p
=
\mathcal N(\boldsymbol\mu_p,\Sigma),
\]
where, as above, \(q\) and \(p\) are the joint posteriors over
\(\bz=\bz_{1:N-1}\) and \(\Sigma\) is the corresponding joint covariance, then,
with \(\boldsymbol\delta=\boldsymbol\mu_q-\boldsymbol\mu_p\), the log-density
ratio \(r\) is affine in \(\bz\), and
\[
D_{\mathrm{KL}}(q\|p)
=
\frac{1}{2}
\boldsymbol\delta^\top
\Sigma^{-1}
\boldsymbol\delta,
\qquad
\operatorname{Var}_q(r)
=
\boldsymbol\delta^\top
\Sigma^{-1}
\boldsymbol\delta.
\]
Hence,
\begin{equation}
\operatorname{Var}_q(r)
=
2D_{\mathrm{KL}}(q\|p),
\qquad
\|b\|
\leq
(S+G_q)
\sqrt{
2D_{\mathrm{KL}}(q\|p)
}.
\label{eqn:appendix-gaussian-total-bound}
\end{equation}

For a locally linear Gaussian bridge at node \(n\), the exact posterior
covariance is
\[
\Sigma_{p,n}
=
\sigma^2
\left(
I+\mathbf J_{r,n}^\top \mathbf J_{r,n}
\right)^{-1},
\]
where \(\mathbf J_{r,n}\) is the right-block Jacobian entering the backward
correction \(\boldsymbol\mu_n^{\leftarrow}\) of Eq.~\eqref{eqn:posterior-mean}
and \(\sigma^2\) is the per-node variance of the reparameterized factor in
Eq.~\eqref{eqn:midpoint-reparameterization},
whereas the TreeProp half-step approximation uses
\[
\Sigma_{q,n}
=
\frac{\sigma^2}{2}I.
\]
They coincide under the orthogonal-Jacobian condition
\[
\mathbf J_{r,n}^\top \mathbf J_{r,n}=I,
\]
equivalently \(\mathbf J_{r,n}\mathbf J_{r,n}^\top=I\) for square blocks. Under the tree factorization
with independent per-node noise, the joint covariances \(\Sigma_p\) and
\(\Sigma_q\) are block diagonal with diagonal entries \(\Sigma_{p,n}\) and
\(\Sigma_{q,n}\), so the equal-covariance hypothesis holds jointly if and only
if it holds at every node. Thus,
Eq.~\eqref{eqn:appendix-gaussian-total-bound} holds exactly under the same
orthogonal-Jacobian condition that makes the half-step inversion exact. When
the normalized residual blocks are close to isometries, the equal-covariance
specialization may provide a useful local approximation, but the general
variance bound in Eq.~\eqref{eqn:appendix-total-bias-bound} remains the
rigorous statement.

It is worth being precise about which of the two distributions the Gaussian
hypothesis constrains. The variational factor \(q_\phi\) is exactly Gaussian by
construction, also for nonlinear blocks: it is defined by
Eq.~\eqref{eqn:midpoint-reparameterization} and sampled exactly by
reparameterization, so block nonlinearity does not make \(q\) approximate. What
fails is the corresponding statement for \(p\). The equal-covariance
specialization is therefore a hypothesis about \(p\), and holds locally in the
linear-Gaussian regime.

If \(q_{\phi}\) admits a differentiable reparameterization and
differentiation may be interchanged with the corresponding expectation, the
score-form expression above and the pathwise derivative used in the
implementation have the same expectation, although their Monte Carlo
estimators need not coincide samplewise. Consequently, using one Monte Carlo
sample affects estimator variance, not the expected gradient characterized
above. Moreover, the active construction paths remain tree-local, so
differentiating through them preserves the
\(\mathcal O(\log N)\) backward span of
Theorem~\ref{thm:logarithmic-learning-path}.

\paragraph{Descent of the original objective.}

Let
\[
g_t
=
\nabla_\theta\mathcal L(\theta_t),
\qquad
\widehat g_t
=
\widetilde{\nabla}_\theta\mathcal F(\theta_t,\phi_t),
\qquad
b_t
=
\widehat g_t-g_t,
\]
and consider the update
\[
\theta_{t+1}
=
\theta_t
-
\eta_t\widehat g_t,
\]
with step size \(\eta_t>0\).
The remaining components of \(\phi\), namely the auxiliary tree-block
parameters, are updated concurrently by their own gradients; the analysis below
treats the resulting trajectory \(\phi_t\) as given and bounds the effect of the
induced bias on \(\mathcal L(\theta_t)\) pointwise along it.
Assume that \(\mathcal L\) is \(L_s\)-smooth, i.e., its gradient is
\(L_s\)-Lipschitz. No convexity assumption is required; the standard descent
lemma holds for smooth nonconvex objectives and gives
\begin{equation}
\mathcal L(\theta_{t+1})
\leq
\mathcal L(\theta_t)
-
\eta_t
\langle
g_t,\widehat g_t
\rangle
+
\frac{L_s}{2}
\eta_t^2
\|\widehat g_t\|^2.
\label{eqn:appendix-descent-lemma}
\end{equation}
Thus, if
\[
\langle
g_t,\widehat g_t
\rangle
>0,
\]
strict descent holds whenever
\begin{equation}
0
<
\eta_t
<
\frac{
2
\langle
g_t,\widehat g_t
\rangle
}{
L_s
\|\widehat g_t\|^2
}.
\label{eqn:appendix-step-size}
\end{equation}

Since \(\widehat g_t=g_t+b_t\),
\[
\begin{aligned}
\langle
g_t,\widehat g_t
\rangle
&=
\|g_t\|^2
+
\langle
g_t,b_t
\rangle
\\
&\geq
\|g_t\|^2
-
\left|
\langle
g_t,b_t
\rangle
\right|
\\
&\geq
\|g_t\|^2
-
\|g_t\|
\|b_t\|
\\
&=
\|g_t\|
\left(
\|g_t\|
-
\|b_t\|
\right),
\end{aligned}
\]
since
\[
\langle
g_t,b_t
\rangle
\geq
-
\left|
\langle
g_t,b_t
\rangle
\right|
\geq
-\|g_t\|
\|b_t\|
\]
by Cauchy--Schwarz. The lower bound corresponds to the worst case in which
the bias is anti-aligned with the true gradient. Therefore,
\begin{equation}
\|b_t\|
<
\|g_t\|
\label{eqn:appendix-bias-descent-condition}
\end{equation}
is sufficient (but not necessary) for positive alignment.

Evaluating \(q\), \(p\), \(r\), \(S\), and \(G_q\) at
\((\theta_t,\phi_t)\), Eq.~\eqref{eqn:appendix-total-bias-bound} shows that it
is sufficient that
\begin{equation}
S
\sqrt{
2D_{\mathrm{KL}}(q\|p)
}
+
G_q
\sqrt{
\operatorname{Var}_q(r)
}
<
\left\|
\nabla_\theta
\mathcal L(\theta_t)
\right\|.
\label{eqn:appendix-general-descent-condition}
\end{equation}
The local-marginal version follows by replacing the first term with the first
term of Eq.~\eqref{eqn:appendix-total-bias-local}.

If \(q\) is held fixed under \(\widetilde{\nabla}_\theta\), then \(S_q=0\)
and hence \(G_q=0\). Using
Eq.~\eqref{eqn:appendix-global-score-bound}, we recover
\begin{equation}
D_{\mathrm{KL}}(q\|p)
<
\frac{
\|
\nabla_\theta
\mathcal L(\theta_t)
\|^2
}{
2S_{\mathrm{glob}}^2
}.
\label{eqn:appendix-fixed-q-descent}
\end{equation}
Under \(S_n\leq S_0\), the more conservative denominator is
\(2NS_0^2\). These are exactly the original fixed-\(q\) conditions.

Finally, in the equal-covariance Gaussian regime,
\begin{equation}
D_{\mathrm{KL}}(q\|p)
<
\frac{
\|
\nabla_\theta
\mathcal L(\theta_t)
\|^2
}{
2(S+G_q)^2
}
\label{eqn:appendix-gaussian-descent}
\end{equation}
is sufficient. All of these conditions are sufficient rather than necessary
and become stricter near stationary points, where the true gradient norm is
small.

The auxiliary parameters do not drop out of these conditions. They enter
through \(D_{\mathrm{KL}}(q_\phi\|p_\theta)\) and through
\(G_q=(\mathbb E_q\|\widetilde{\nabla}_\theta\log q_\phi\|^2)^{1/2}\), both of
which are evaluated at the current \(\phi_t\); no redefinition of \(\theta\) has
taken place. The conditions are therefore joint conditions on \(\phi\), and
training the auxiliary components to reduce the per-node consistency losses is
exactly what makes them easier to satisfy. The case \(S_q=0\) above is the
degenerate limit in which \(q\) is held fixed under
\(\widetilde{\nabla}_\theta\). Figure~\ref{fig:local_kl} measures the
\(\phi\)-dependent divergence directly along a training run.

\begin{figure}[t]
\centering
\includegraphics[width=\columnwidth]{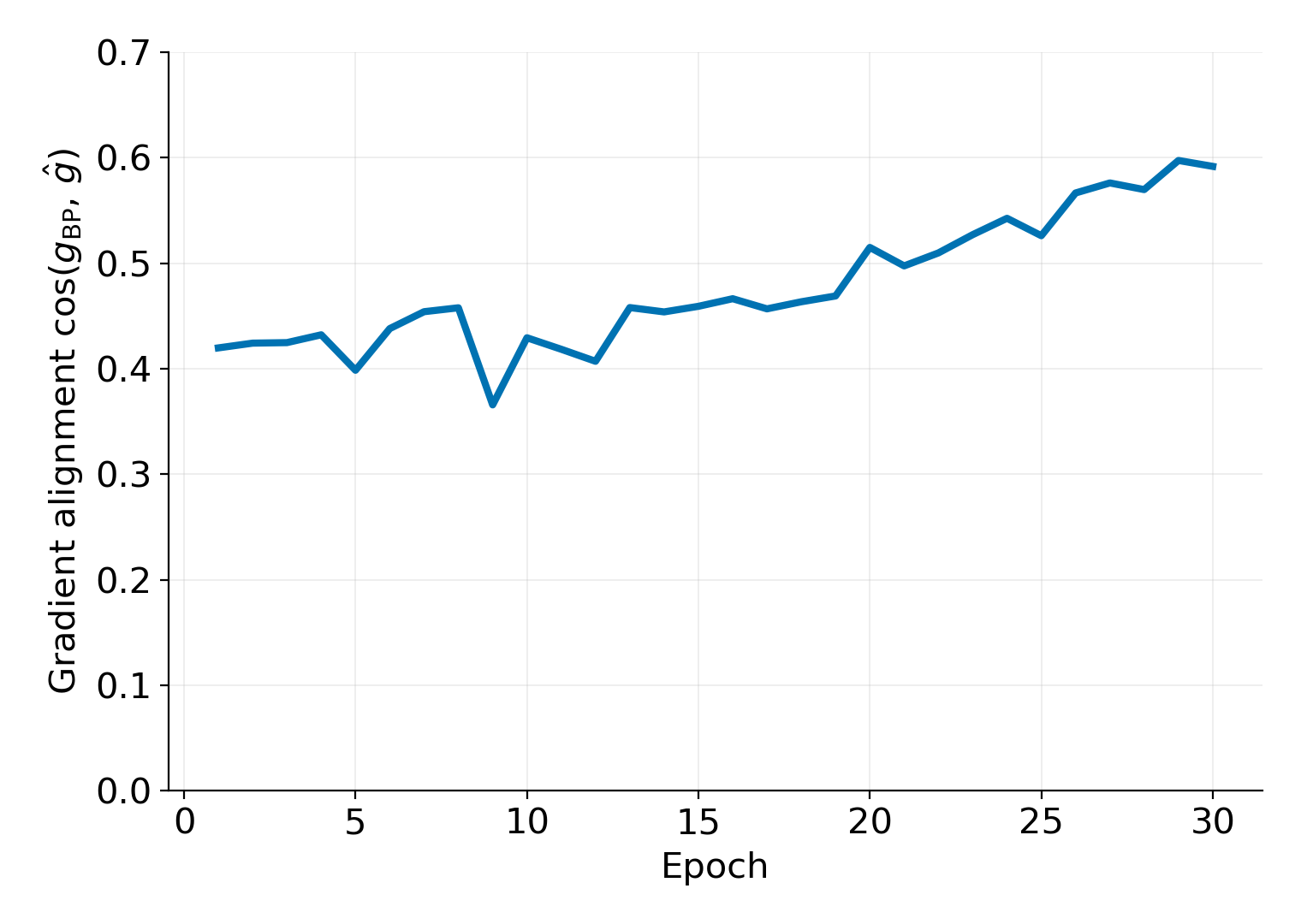}
\caption{Alignment $\cos(g_{\mathrm{BP}}, \hat{g})$ between the true gradient
and the TreeProp update (same run as
Figure~\ref{fig:local_kl}).The curve is a single
alignment for the deployed chain as a whole, not a per-node quantity. The update
stays in the descent half-space throughout and aligns better as the local losses
vanish.}
\label{fig:grad_align}
\end{figure}

Figures~\ref{fig:local_kl} and~\ref{fig:grad_align} provide empirical validation for the analysis on a
depth-3 model trained on WikiText-103 for 30 epochs. TreeProp never computes the end-to-end
gradient: it descends a surrogate of local objectives, and inherits progress
on the true objective only to the extent that its update stays correlated
with the true gradient. The analysis makes this quantitative (the update
decomposes as $\hat{g} = g_{\mathrm{BP}} + b$ with the bias $b$ controlled by
the per-node KL consistency losses) so measuring both along a real
training trajectory tests the analysis end-to-end.
Once per epoch we compute,
on a held-out probe batch, the exact end-to-end gradient $g_{\mathrm{BP}}$ and
the TreeProp update direction $\hat{g}$ on the same parameters (gradients are with respect to the full weight matrix). Here $g_{\mathrm{BP}}$ is obtained by backpropagating the deployed model's
cross-entropy through all layers, and $\hat{g}$ is the gradient of the
composite local objective TreeProp actually applies; both are restricted to
the parameters BP touches, flattened, and compared by cosine similarity.
The probe is trajectory-neutral: optimizer and RNG state are saved and
restored, so the measurement does not perturb the run it measures. Both gradients are flattened and concatenated over
the parameters of the deployed depth-$0$ chain, that is, every parameter that
end-to-end backpropagation reaches; the tree-only auxiliary blocks receive no
backpropagated gradient and are excluded. 
Figure~\ref{fig:local_kl} shows the per-node local KL consistency losses, the
quantities that control the bias term $b$: all nodes converge to near zero by the end of training. The divergence $D_{\mathrm{KL}}
\left(
q_{\phi}(\bz\mid\bx,\by)
\middle\|
p_\theta(\bz\mid\bx,\by)
\right)$ is exactly the quantity whose gradient is the bias, since
$\mathcal F=\mathcal L+D_{\mathrm{KL}}$ gives
$b=\widetilde{\nabla}_\theta D_{\mathrm{KL}}(q_\phi\|p_\theta)$, and its decay
bounds $\|b\|$ through Eq.~\eqref{eqn:appendix-total-bias-bound}. Figure~\ref{fig:grad_align} shows the resulting
alignment: $\cos(g_{\mathrm{BP}}, \hat{g})$ remains strictly positive at every
epoch (the update never leaves the descent half-space, i.e., every step computed
from purely local quantities still decreases the global loss to first order) and \emph{improves}
from ${\approx}0.42$ to ${\approx}0.59$ as the local losses vanish. We note that the alignment tracks the local-loss decay exactly as the analysis
predicts.

\subsection{Proof of Theorem~\ref{thm:logarithmic-learning-path}}
\label{app:proof-logarithmic-learning-path}

We show that both the forward construction of the tree-structured
representations and the corresponding backward gradient propagation have critical-path depth \(\mathcal{O}(\log N)\). 
The forward computation proceeds level by level from the root toward the lowest tree level, while the backward computation traverses the same tree levels in reverse order.

\paragraph{Forward computation.}
We first consider \(N=2^D\). The construction starts from the root interval
\[
\mathcal{B}_1=\{(0,N)\},
\]
The root midpoint \(n=N/2\) is introduced at the first tree level.
Since the boundary representations \(\bz_0=\bx\) and \(\bz_N=\by\) are available during training, the root representation \(\bz_{N/2}\) can be estimated at the
first tree level from
\(q_{\phi_{N/2}}(\bz_{N/2}\mid\bx,\by)\) using the tree-structured variational factor estimation procedure defined in
Eqs.~\eqref{eqn:midpoint-reparameterization}.

This introduces the root representation and splits the root interval into
\[
\left(0,\frac{N}{2}\right)
\qquad\text{and}\qquad
\left(\frac{N}{2},N\right),
\]
each of length \(N/2\).

At the second tree level, the two child intervals are processed
simultaneously. Their boundary representations are already available, so
their midpoint representations can be estimated independently. This produces
four child intervals, each of length \(N/4\).

Continuing recursively, tree level \(s\) contains
\[
2^{s-1}
\]
active intervals, each of length
\[
\frac{N}{2^{s-1}}.
\]
For every interval \([l,r]\in\mathcal{B}_s\), the midpoint representation
\[
\bz_n
\sim
q_{\phi_n}
\left(
\bz_n\mid\bz_l,\bz_r
\right)
\]
depends only on the boundary representations \(\bz_l\) and \(\bz_r\),
which are available before level \(s\) is evaluated. Midpoints introduced
from distinct active intervals at the same level do not depend on one
another. Hence, all midpoint representations at a fixed tree level can be
estimated in parallel.

At the lowest tree level \(s=D\), the active intervals have length
\[
\frac{N}{2^{D-1}}
=
2.
\]
Each such interval contains exactly one remaining intermediate
representation. Estimating its midpoint splits the interval into two
unit-length intervals, which contain no intermediate representations and
require no further computation. Therefore, the construction terminates after
exactly
\[
D=\log_2N
\]
tree levels.

Since the \(D\) tree levels are sequentially dependent, while all midpoint
representations within each level are evaluated in parallel, the forward
critical-path depth is
\begin{equation}
T_{\mathrm{forward}}
=
D
=
\log_2N.
\label{eqn:appendix-forward-depth}
\end{equation}

\paragraph{Backward computation.}
Let
\[
\mathcal{R}_s
=
\left\{
(l,n,r):
(l,r)\in\mathcal{B}_s,\;
n=
\left\lfloor\frac{l+r}{2}\right\rfloor,\;
r-l\geq 2
\right\}
\]
denote the set of interval--midpoint relations introduced at tree level
\(s\). Each relation \((l,n,r)\in\mathcal{R}_s\) introduces the
representation
\[
\bz_n\sim q_{\phi_n}(\bz_n\mid\bz_l,\bz_r)
\]
and has an associated node objective \(\mathcal{F}_{l,n,r}\), as defined
in Appendix~\ref{app:nodewise-losses}. The complete objective is
\[
\mathcal{F}_{\mathrm{tree}}
=
\sum_{s=1}^{D}
\sum_{(l,n,r)\in\mathcal{R}_s}
\mathcal{F}_{l,n,r}.
\]

Let
\[
\mathcal{C}_s
=
\left\{
\bz_n:
(l,n,r)\in\mathcal{R}_s
\right\}
\]
denote the set of representations introduced at tree level \(s\).
Thus, \(\mathcal{C}_1\) contains the root representation, while
\(\mathcal{C}_D\) contains the representations at the lowest tree level.

By construction of the tree, each representation in \(\mathcal{C}_s\)
depends only on boundary representations available from earlier tree levels,
with no computational dependencies between representations introduced at
the same level. As shown in the forward-computation derivation, the longest
forward dependency path therefore contains at most \(D\)
representation-construction steps.

Reverse-mode differentiation traverses these dependencies in reverse. Let
\[
\boldsymbol{\lambda}_n
=
\nabla_{\bz_n}\mathcal{F}_{\mathrm{tree}}
\]
denote the adjoint of \(\bz_n\). For an interval relation \((l,n,r)\),
differentiating the sampled representation \(\bz_n\)
produces gradient contributions to its boundary representations
\(\bz_l\) and \(\bz_r\), together with direct gradient contributions from
the associated node objective.

Reversing the dependency directions preserves the longest path length.
Since nodes at the same tree level have no mutual dependencies, their
adjoints and parameter-gradient contributions,
\[
\left\{
\boldsymbol{\lambda}_n,\,
\boldsymbol{g}_{l,n,r}
\right\}_{(l,n,r)\in\mathcal{R}_s},
\]
can be computed in parallel, where \(\boldsymbol{g}_{l,n,r}\)
denotes the parameter-gradient contribution associated with the node
objective \(\mathcal{F}_{l,n,r}\).

The backward traversal processes the levels in the order
\[
\mathcal{C}_D,
\mathcal{C}_{D-1},
\ldots,
\mathcal{C}_1,
\]
requiring one reverse step per tree level to accumulate the gradients in that level. 
Hence,
\[
T_{\mathrm{backward}}
\leq
D
=
\left\lceil\log_2 N\right\rceil.
\]

The forward traversal likewise requires at most one parallel step per tree
level, giving
\[
T_{\mathrm{forward}}
\leq D.
\]
Consequently,
\[
T_{\mathrm{forward\text{--}backward}}
\leq
2D
=
2\left\lceil\log_2 N\right\rceil
=
\mathcal{O}(\log N).
\]

The computation performs \(\mathcal{O}(N)\) total representation and node
objective evaluations. The resulting objective and parameter-gradient
contributions can be combined using parallel reductions of depth
\(\mathcal{O}(\log N)\), which does not change the asymptotic
forward-backward span.
\qed

\section{Extensions and Additional Analysis}
\label{app:experimental-results}

\subsection{Recurrent Sequence Modeling}
\label{app:recurrent_TreeProp}

Although the main experiments in this work focus on feedforward and Transformer-style architectures, the TreeProp formulation can also be adapted to recurrent sequence modeling.
This is natural because recurrent architectures define Markovian transitions over hidden states, which align with the conditional factorization in Theorem~\ref{thm:chains-to-trees}.
In this setting, the hierarchy is interpreted over time indices, and each node summarizes a block of past inputs.
However, the recurrent cell deployed at inference must be causal: the lowest-depth cell consumes only the previous hidden state and the current input, yielding a standard autoregressive recurrence. To ensure this, we obtain a causal variant through a single change to how the inter-node message is formed. A node at depth $d(n)$ folds its left input block, the $2^{d(n)}$ symbols $\bx_{n-2^{d(n)}},\dots,\bx_{n-1}$, a span that doubles with depth, into a state passed to its children, 
\begin{equation}
\bh_n = f_{d(n)}\big(\bh_{n-2^{d(n)}}, \bx_{n-2^{d(n)}:n-1}\big)
\end{equation}
propagated from the left parent only. The future (right) parent enters solely as a training target for local learning. Each depth $d$ therefore has its own cell $f_d$ sharing a single hidden width $H$, and cells higher in the tree have larger input sizes in proportion to the longer input spans ($2^{d}$ symbols) they summarise. This construction is causal with full coverage: every node's receptive field is exactly the prefix $\bx_0,\dots,\bx_n$, it depends on all past inputs and no future ones, and every depth sees the full input, which is necessary for the per depth RNNs to train. 
Thus, the tree provides a parallel training structure, while inference follows the usual causal recurrent rollout of the single shared depth-0 cell $f_0$.

Inside the tree the finest chain is teacher-forced (each node reads the clean predecessor from its left parent), but at inference the cell consumes its own output, $\bh_t = f_0(\bh_{t-1}, \bx_t)$, so errors compound over the $N$ steps. The deployed states are then drawn from a self-generated distribution unlike the clean training one, an exposure-bias mismatch that leaves the tree readouts accurate while the rolled chain degrades with length, collapsing at the largest depths.
To address the mismatch between tree-based training and autoregressive rollout at inference, we experimented with a parallel consistency stage that exposes the recurrent cell to self-generated predecessor states without requiring a full sequential rollout. 
The tree pass first produces a clean target state $\bar{\mathbf{h}}_t$ for each position $t$, which is detached from the computational graph. We initialize the consistency states using these tree-generated targets: \begin{align} 
\tilde{\mathbf{h}}^{(0)}_t &= \bar{\mathbf{h}}_t . \label{eq:consistency_init} 
\end{align} 
For each consistency sweep $i=1,\ldots,K$, every position is recomputed using the previous position's state from the preceding sweep: \begin{align} 
\tilde{\mathbf{h}}^{(i)}_t &= f_0\!\left( \tilde{\mathbf{h}}^{(i-1)}_{t-1}, \mathbf{x}_t \right), \qquad i=1,\ldots,K . \label{eq:consistency_sweep} 
\end{align} 
Equivalently, this corresponds to applying the deployed recurrent transition \[ p_{\theta}(\mathbf{h}_t \mid \mathbf{h}_{t-1},\mathbf{x}_t) \] using predecessor states generated by the model itself in the previous sweep. 
Importantly, within a fixed sweep $i$, all states $\tilde{\mathbf{h}}^{(i-1)}_{t-1}$ are treated as fixed values from the previous sweep. Therefore, the updates over all positions $t$ are conditionally independent and can be computed in parallel. 
After $K$ sweeps, we match the corrected states to the clean tree targets using the consistency objective 
\begin{align} \mathcal{L}_{\mathrm{cons}} &= \sum_t \left\| \tilde{\mathbf{h}}^{(K)}_t - \bar{\mathbf{h}}_t \right\|_2^2 . \label{eq:consistency_loss} 
\end{align} 
This trains the deployed recurrent cell $f_0$ on states closer to the
self-generated distribution encountered during autoregressive inference, while avoiding a full sequential rollout during training. 
As a result, the recurrent cell is encouraged to remain stable under the state distribution encountered during deployment. 
The motivation for this correction is related to DAgger, which addresses
sequential covariate shift by training a learner on states induced by its own
policy rather than only on states from expert trajectories
\citep{ross2011reduction}. In TreeProp-TT, an analogous shift occurs in
hidden-state space: the detached tree trajectory
$(\bar{\mathbf{h}}_{1:N})$ provides reference targets, while the deployed cell
(f\_0) is evaluated on progressively self-generated predecessor states
$(\tilde{\mathbf{h}}^{(i-1)}_{t-1})$. This analogy is not exact, since our
procedure uses neither an external expert nor explicit dataset aggregation,
and the reference target $(\bar{\mathbf{h}}\_t)$ is not recomputed at each
perturbed state. Computationally, the repeated parallel sweeps are related to
Parallel Scheduled Sampling, which exposes autoregressive models to
model-generated histories through successive parallel passes
\citep{duckworth2019parallel}. Our procedure instead applies the passes to
continuous recurrent states and supervises the corrected states using
tree-derived targets.

The $logN$ span is preserved for $K \le logN$, which is true in practice as $N$ grows. To avoid propagating errors caused by inversion (since all cells in a chain share weights, any error propagates), we avoid it altogether by learning the forward direction of the usually inverted right model and not performing the fusion step. Since this is still a constant multiple, the parallel $O(log N)$ span is preserved.

\subsubsection{Experimental Setup}

We evaluate the recurrent variant on sequential MNIST and permuted-sequential
MNIST; the architecture, input construction, and hyperparameters are specified
in Appendix~\ref{app:smnist_details}. In brief, at tree depth $D$ the $784$
pixels are presented as $N=2^{D}$ steps, the deployed model is the single
depth-$0$ cell unrolled over that sequence, and the backpropagation-through-time
(BPTT) baseline matches it in cell type, hidden width, and parameter count, with
optimizer, input preprocessing, and training schedule shared between the two
methods within each task. Permuted-sequential MNIST applies one fixed random
permutation to the $784$ pixels, identical for every image and for both methods,
which destroys spatial locality and forces integration across the whole
sequence. For the results in Table~\ref{tab:recurrent_results}, $K=0$ and $K=4$
were used at $D=6$ and $D=8$ respectively for permuted-sequential MNIST, and
$K=0$ for sequential MNIST.

\subsubsection{Results}

Across both sequential MNIST and permuted-sequential MNIST, TreeProp-TT remains competitive with the corresponding BPTT baselines for both GRU and gateless Elman-style recurrent cells, as can be seen in Table \ref{tab:TreeProp-all}. For the GRU cell, the deployed TreeProp sequential chain matches or slightly improves over BPTT for shallower and moderate depths, while remaining close to BPTT even as the rollout length increases. 
On permuted-sequential MNIST, where long-range integration is more important, the TreeProp tree readout remains highly stable across depths,
achieving around $98.5$--$98.7\,\%$ accuracy, while the deployed sequential chain gradually degrades at the largest depths. It should be noted that the TreeProp seq. accuracy is higher than BPTT for depths 1 through 7 on permuted-sequential MNIST, and comparable for depth 8, showing TreeProp's ability to more effectively capture long range dependencies.

The gateless Elman-style cell shows a similar trend. TreeProp remains competitive with BPTT on both sequential and permuted-sequential MNIST. The tree readout is especially stable across depths, whereas the deployed sequential chain shows a mild degradation for longer rollouts. 
These results suggest that the hierarchical training tree learns strong recurrent representations, and that the consistency correction helps
bridge the gap between tree-based training and causal sequential deployment. We note that gateless Elman chain is best left uncorrected ($K{=}0$),
    as the consistency iteration destabilises it. 

Figure \ref{fig:speedup-gru-mnist} also demonstrates the speedup achieved at different depths by TreeProp is significant as $D$ grows, achieving upto a 3.6x speedup at depth 8. TreeProp is slower than BPTT at
shallow depth, where the per-step launch overhead dominates and the tree has few levels to
parallelise, but crosses unity near $D{=}6$ and grows to ${\sim}3.6\times$ at $D{=}8$, as the
$\mathcal{O}(\log N)$ training span increasingly outpaces the baseline's $\mathcal{O}(N)$ unroll, at
no accuracy cost. We expect this trend to continue as depth is increased further, following the expected $C.N/logN$ speedup curve.

\subsection{Training Speedup on Sequential MNIST for TreeProp-TT}
\label{app-rnn-speedup}

\begin{figure} [!htbp]
    \centering
    \includegraphics[width=1.0\linewidth]{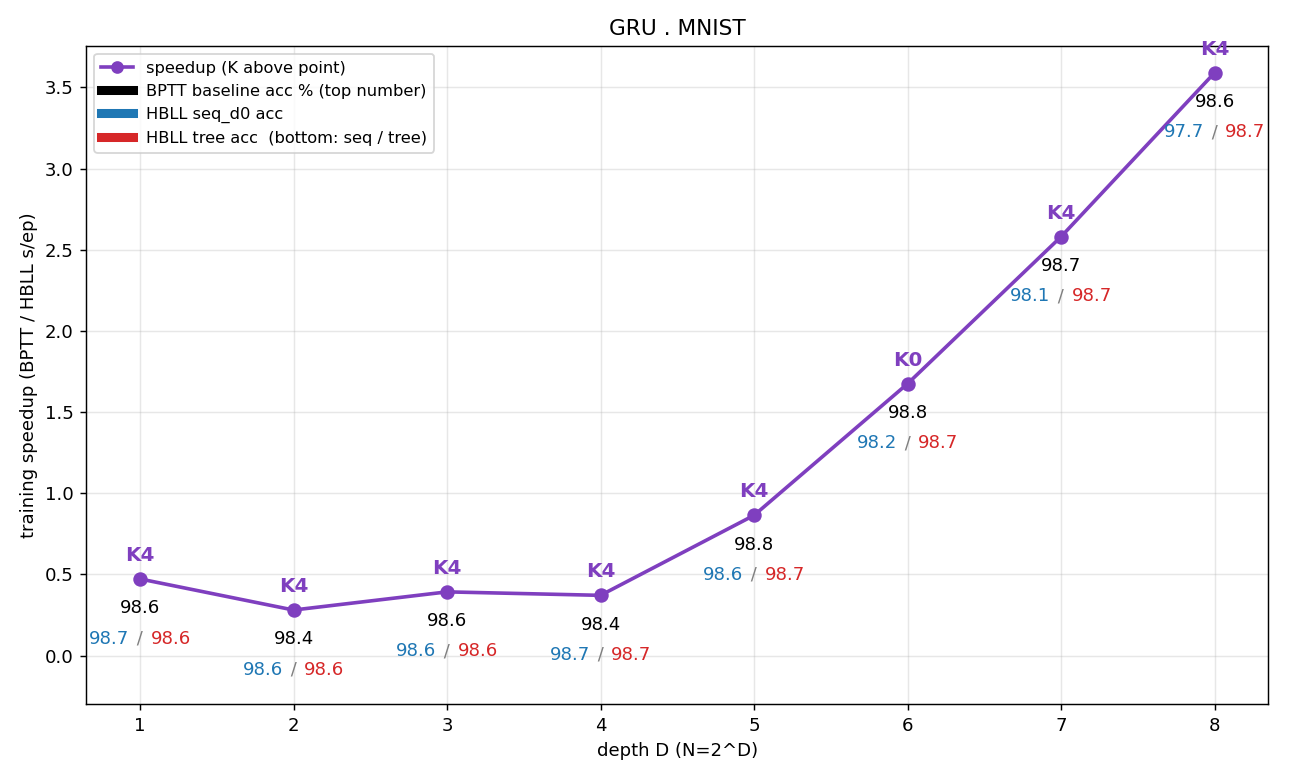}
    \caption{Per-epoch training speedup of TreeProp-TT over a matched BPTT baseline on sequential MNIST using a GRU with hidden width \(H=2048\). Tree depths \(D=1,\dots,8\) correspond to sequence lengths \(N=2^D\). Values above \(1\) indicate faster TreeProp training. Labels above the points show the selected consistency-iteration count \(K\); annotations below report BPTT and TreeProp sequential/tree accuracies.}
    \label{fig:speedup-gru-mnist}
\end{figure}

Figure~\ref{fig:speedup-gru-mnist} shows that TreeProp-TT is slower than BPTT at small hierarchy depths, where the overhead of constructing the tree and performing the consistency correction dominates. The crossover occurs around \(D=6\), after which the increasing parallelism of the hierarchical training procedure outweighs this overhead. 
At \(D=8\), corresponding to a sequence length of \(N=256\), TreeProp reaches approximately \(3.6\times\) per-epoch speedup while maintaining comparable deployed-sequence accuracy and a stable tree-readout accuracy. We note these numbers are obtained on limited compute (a single GH200 GPU) and the true estimated speedup can be achieved given sufficient parallel compute. 

\begin{table*}[t]
    \centering
    \small
    \begin{tabular}{ccccc|ccc}
        \toprule
        \multicolumn{5}{c}{Configuration} & \multicolumn{3}{c}{Accuracy (\%)} \\
        \cmidrule(lr){1-5} \cmidrule(lr){6-8}
        Dataset & Cell & Depth $D$ & Length $N$ & Best $K$ & TreeProp seq. & TreeProp tree & BPTT \\
        \midrule
        MNIST & GRU & $1$ & $2$   & $4$ & $98.7$ & $98.6$ & $98.6$ \\
              &     & $2$ & $4$   & $4$ & $98.6$ & $98.6$ & $98.4$ \\
              &     & $3$ & $8$   & $4$ & $98.6$ & $98.6$ & $98.6$ \\
              &     & $4$ & $16$  & $4$ & $98.7$ & $98.7$ & $98.4$ \\
              &     & $5$ & $32$  & $4$ & $98.6$ & $98.7$ & $98.8$ \\
              &     & $6$ & $64$  & $0$ & $98.2$ & $98.7$ & $98.8$ \\
              &     & $7$ & $128$ & $4$ & $98.1$ & $98.7$ & $98.7$ \\
              &     & $8$ & $256$ & $4$ & $97.7$ & $98.7$ & $98.6$ \\
        \cmidrule(lr){2-8}
              & Elman RNN& $1$ & $2$  & $0$ & $98.7$ & $98.7$ & $98.6$ \\
              &       & $2$ & $4$  & $0$ & $98.4$ & $98.7$ & $98.5$ \\
              &       & $3$ & $8$  & $0$ & $98.4$ & $98.6$ & $98.5$ \\
              &       & $4$ & $16$ & $0$ & $98.4$ & $98.6$ & $98.4$ \\
              &       & $5$ & $32$ & $0$ & $98.5$ & $98.7$ & $98.1$ \\
              &       & $6$ & $64$ & $0$ & $98.0$ & $98.7$ & $97.9$ \\
        \midrule
        psMNIST & GRU & $1$ & $2$   & $4$ & $98.7$ & $98.6$ & $98.6$ \\
                &     & $2$ & $4$   & $0$ & $98.5$ & $98.7$ & $98.4$ \\
                &     & $3$ & $8$   & $4$ & $98.6$ & $98.6$ & $98.4$ \\
                &     & $4$ & $16$  & $4$ & $98.5$ & $98.5$ & $98.1$ \\
                &     & $5$ & $32$  & $4$ & $98.3$ & $98.6$ & $97.6$ \\
                &     & $6$ & $64$  & $4$ & $97.9$ & $98.7$ & $96.8$ \\
                &     & $7$ & $128$ & $0$ & $97.3$ & $98.7$ & $95.9$ \\
                &     & $8$ & $256$ & $4$ & $95.1$ & $98.7$ & $95.3$ \\
        \cmidrule   (lr){2-8}
                & Elman RNN & $1$ & $2$  & $0$ & $98.4$ & $98.4$ & $98.5$ \\
                &       & $2$ & $4$  & $0$ & $98.4$ & $98.4$ & $98.2$ \\
                &       & $3$ & $8$  & $0$ & $98.3$ & $98.4$ & $98.3$ \\
                &       & $4$ & $16$ & $0$ & $98.2$ & $98.5$ & $98.2$ \\
                &       & $5$ & $32$ & $0$ & $98.1$ & $98.4$ & $97.5$ \\
                &       & $6$ & $64$ & $0$ & $97.2$ & $98.4$ & $97.0$ \\
        \bottomrule
    \end{tabular}
    \caption{TreeProp classification accuracy (\%) on sequential and permuted-sequential MNIST, for a GRU
    and a gateless Elman cell at hidden width $H{=}2048$, across tree depths $D$ (sequence length
    $N{=}2^{D}$). TreeProp-TT denotes the proposed recurrent model. The
    TreeProp tree-readout and deployed sequential columns come from the \emph{same} jointly trained
    hierarchical model: the tree readout uses the full $O(\log N)$-span tree formed by starting at the
    first node and following all left models (the in-model ceiling), while seq.\ is the depth-$0$ base
    cell unrolled as an ordinary step-by-step RNN, the path actually deployed at inference. ``Best
    $K$'' is the number of fixed-point (consistency) correction iterations used to train the deployed
    chain ($K{=}0$ denotes no correction). BPTT denotes standard end-to-end backpropagation
    through time, trained separately with the same cell architecture and parameter count as seq.}
    \label{tab:TreeProp-all}
\end{table*}

\subsection{Subnetwork Analysis}
\label{app:subnetwork_analysis}

TreeProp naturally induces a family of subnetworks corresponding to
different paths through the learned hierarchy, as illustrated in
Fig.~\ref{fig:tree-prop}. These subnetworks reuse the auxiliary
parameters introduced for training the tree-structured variational
framework. To enable accurate inference through shorter hierarchical
paths, we include cross-entropy losses at the corresponding intermediate
levels. The associated predictions are already computed during posterior
inference for the rightmost nodes, and the corresponding output heads
share parameters across hierarchy levels. Without the intermediate-level
cross-entropy terms, the training objective primarily optimizes the full
sequential path, as reflected in the main results in
Table~\ref{tab:classification_results}. In the ablation study, we examine
how different combinations of node-level and cross-entropy losses affect
the capabilities of the resulting subnetworks.

To evaluate this property, we consider TreeProp models of depths \(D=3\)
and \(D=4\). The depth-\(3\) model contains activations
\(\mathbf{z}_0,\mathbf{z}_1,\ldots,\mathbf{z}_8\), where
\(\mathbf{z}_0=\mathbf{x}\) denotes the input and
\(\mathbf{z}_8=\mathbf{y}\) denotes the output. In addition to the full
path
\[
\mathbf{x}\rightarrow\mathbf{z}_1\rightarrow\mathbf{z}_2\cdots\rightarrow\mathbf{z}_7\rightarrow\mathbf{y},
\]
we evaluate the following shorter hierarchical paths extracted from the
same jointly trained TreeProp model:
\[
\begin{aligned}
&\mathbf{x}\rightarrow\mathbf{z}_4\rightarrow\mathbf{y},\\
&\mathbf{x}\rightarrow\mathbf{z}_4\rightarrow\mathbf{z}_6
\rightarrow\mathbf{y},\\
&\mathbf{x}\rightarrow\mathbf{z}_4\rightarrow\mathbf{z}_6
\rightarrow\mathbf{z}_7\rightarrow\mathbf{y}.
\end{aligned}
\]
These paths correspond to subnetworks with one, two, and three
Transformer blocks, respectively.

For the depth-\(4\) language model, we consider activations
\(\mathbf{z}_0,\mathbf{z}_1,\ldots,\mathbf{z}_{13}\), where
\(\mathbf{z}_0=\mathbf{x}\) and \(\mathbf{z}_{13}=\mathbf{y}\). Since
this configuration contains an odd number of computational blocks, the
final transition is implemented using an additional Transformer block
followed by the language-modeling head, rather than using the
language-modeling head alone. We evaluate the shorter hierarchical paths
\[
\begin{aligned}
&\mathbf{x}\rightarrow\mathbf{z}_5\rightarrow\mathbf{y},\\
&\mathbf{x}\rightarrow\mathbf{z}_5\rightarrow\mathbf{z}_9
\rightarrow\mathbf{y},\\
&\mathbf{x}\rightarrow\mathbf{z}_5\rightarrow\mathbf{z}_9
\rightarrow\mathbf{z}_{11}\rightarrow\mathbf{y},\\
&\mathbf{x}\rightarrow\mathbf{z}_5\rightarrow\mathbf{z}_9
\rightarrow\mathbf{z}_{11}\rightarrow\mathbf{z}_{12}
\rightarrow\mathbf{y},
\end{aligned}
\]
together with the full sequential path
\[
\mathbf{x}\rightarrow\mathbf{z}_1\rightarrow\mathbf{z}_2
\rightarrow\cdots\rightarrow\mathbf{z}_{12}\rightarrow\mathbf{y}.
\]

All TreeProp subnetworks share parameters from the corresponding jointly
trained hierarchical model. For comparison, each BP baseline is trained
separately using end-to-end backpropagation with the same effective
number of Transformer blocks. For CIFAR-10, CIFAR-100, and TinyImageNet,
the prediction head is a classification head, whereas for WikiText-103
it is the language-modeling head.
\begin{table*}[t]
\centering
\small
\setlength{\tabcolsep}{4pt}
\begin{tabular}{lc cc cc cc}
\toprule
& &
\multicolumn{2}{c}{CIFAR-10}
& \multicolumn{2}{c}{CIFAR-100}
& \multicolumn{2}{c}{TinyImageNet} \\
\cmidrule(lr){3-4}
\cmidrule(lr){5-6}
\cmidrule(lr){7-8}
TreeProp path
& \shortstack{Transformer\\blocks}
& BP & TreeProp
& BP & TreeProp
& BP & TreeProp \\
\midrule

$\bx\!\rightarrow\!\bz_4\!\rightarrow\!\by$
& 1
& $75.79 \pm 0.50$ & $78.60 \pm 0.21$
& $54.21 \pm 0.24$ & $50.54 \pm 0.71$
& $38.38 \pm 0.26$ & $35.36 \pm 0.30$ \\

$\bx\!\rightarrow\!\bz_4\!\rightarrow\!\bz_6\!\rightarrow\!\by$
& 2
& $83.11 \pm 0.08$ & $85.91 \pm 0.06$
& $54.52 \pm 0.52$ & $58.18 \pm 0.40$
& $43.38 \pm 0.52$ & $41.30 \pm 0.21$ \\

$\bx\!\rightarrow\!\bz_4\!\rightarrow\!\bz_6
\!\rightarrow\!\bz_7\!\rightarrow\!\by$
& 3
& $85.55 \pm 0.55$ & $87.36 \pm 0.11$
& $58.57 \pm 0.44$ & $59.04 \pm 0.40$
& $43.88 \pm 0.15$ & $41.80 \pm 0.15$ \\

$\bx\!\rightarrow\!\bz_1\!\rightarrow\!\cdots\!\rightarrow\!\by$
& 7
& $88.62 \pm 0.84$ & $86.83 \pm 0.08$
& $59.12 \pm 0.48$ & $58.17 \pm 0.40$
& $43.91 \pm 0.15$ & $41.41 \pm 0.21$ \\

\bottomrule
\end{tabular}

\caption{Top-\(1\) classification accuracy of subnetworks extracted from
the same jointly trained TreeProp hierarchy. Each row corresponds to a
different inference path and effective number of Transformer blocks.
BP results are obtained from separately trained end-to-end models with
the corresponding effective depth.}
\label{tab:classification_subnetworks}
\end{table*}

For CIFAR-10, CIFAR-100 and TinyImageNet, Table~\ref{tab:classification_subnetworks} show that the subnetworks sampled from a single jointly trained TreeProp hierarchy achieve performance comparable to separately trained BP models across different effective depths. 
This indicates that the same TreeProp model supports multiple inference paths without requiring a separate model to be trained for each number of Transformer blocks.

\begin{table*}[t]
\centering
\small
\setlength{\tabcolsep}{5pt}
\begin{tabular}{clc|cc}
\toprule
Tree depth
& TreeProp path
& \shortstack{Transformer\\blocks}
& \multicolumn{2}{c}{Perplexity} \\
\cmidrule(lr){4-5}
& & & BP & TreeProp \\
\midrule

\multirow{4}{*}{\(D=3\)}
& \(\mathbf{x}\rightarrow\mathbf{z}_4\rightarrow\mathbf{y}\)
& 1
& \(32.42 \pm 0.12\)
& \(34.08 \pm 0.13\) \\

& \(\mathbf{x}\rightarrow\mathbf{z}_4\rightarrow
   \mathbf{z}_6\rightarrow\mathbf{y}\)
& 2
& \(23.77 \pm 0.02\)
& \(24.04 \pm 0.05\) \\

& \(\mathbf{x}\rightarrow\mathbf{z}_4\rightarrow
   \mathbf{z}_6\rightarrow\mathbf{z}_7\rightarrow\mathbf{y}\)
& 3
& \(22.15 \pm 0.06\)
& \(22.04 \pm 0.05\) \\

& \(\mathbf{x}\rightarrow\mathbf{z}_1\rightarrow
   \mathbf{z}_2\rightarrow\cdots\rightarrow\mathbf{y}\)
& 7
& \(19.75 \pm 0.07\)
& \(22.16 \pm 0.04\) \\

\midrule

\multirow{5}{*}{\(D=4\)}
& \(\mathbf{x}\rightarrow\mathbf{z}_5\rightarrow\mathbf{y}\)
& 2
& \(23.77 \pm 0.02\)
& \(22.46 \pm 0.04\) \\

& \(\mathbf{x}\rightarrow\mathbf{z}_5\rightarrow
   \mathbf{z}_9\rightarrow\mathbf{y}\)
& 3
& \(22.15 \pm 0.06\)
& \(19.69 \pm 0.04\) \\

& \(\mathbf{x}\rightarrow\mathbf{z}_5\rightarrow
   \mathbf{z}_9\rightarrow\mathbf{z}_{11}\rightarrow\mathbf{y}\)
& 4
& \(21.10 \pm 0.20\)
& \(18.45 \pm 0.01\) \\

& \(\mathbf{x}\rightarrow\mathbf{z}_5\rightarrow
   \mathbf{z}_9\rightarrow\mathbf{z}_{11}\rightarrow
   \mathbf{z}_{12}\rightarrow\mathbf{y}\)
& 5
& \(20.00 \pm 0.28\)
& \(17.76 \pm 0.01\) \\

& \(\mathbf{x}\rightarrow\mathbf{z}_1\rightarrow
   \mathbf{z}_2\rightarrow\cdots\rightarrow
   \mathbf{z}_{12}\rightarrow\mathbf{y}\)
& 12
& \(16.43 \pm 0.04\)
& \(19.49 \pm 0.08\) \\

\bottomrule
\end{tabular}

\caption{Subnetwork evaluation for TreeProp on WikiText-103. Perplexity
is reported for subnetworks obtained from different hierarchical paths
of the same trained TreeProp model. BP results correspond to separately
trained end-to-end baselines with the same effective number of
Transformer blocks and are shared across hierarchy depths. Lower perplexity
is better.}
\label{tab:wikitext103_subnetwork}
\end{table*}

Table~\ref{tab:wikitext103_subnetwork} shows that the shorter TreeProp
language models remain reasonably close to their separately trained BP
counterparts. For this subnetwork comparison, we use a learning rate of
\(1.5\times10^{-4}\) for the depth-\(3\) model and
\(4\times10^{-4}\) for the depth-\(4\) model. Both configurations use a
global batch size of \(32\), corresponding to \(8\) samples per GPU
across four GPUs.

Overall, these results show that a single jointly trained TreeProp model supports multiple inference paths with different effective depths. 
In contrast, the BP results require a separate end-to-end model to be trained for every evaluated depth.





\section{Additional Experimental Details}
\label{app:experimental-details}
This section provides additional implementation details, hyperparameter
settings, and extended results for the experiments presented in
the Experiments section.

We report the ideal \(T_{\infty}\) speedup in the Experiments section by comparing the combined forward-backward critical-path depths of BP and TreeProp, assuming sufficient parallel compute resources and equal cost for each computational step. 
Here, \(N\) denotes the total number of sequential computational blocks, including the classifier head. 
BP has a combined critical-path depth of \(2N\), accounting for \(N\) sequential steps in both the forward and backward passes. 
For a TreeProp hierarchy of depth \(D=\lceil\log_2 N\rceil\), the corresponding critical-path depth is \(2D\), since all nodes at the same tree level can be evaluated in parallel. The ideal speedup is therefore
\[
S_{\infty}
=
\frac{T_{\infty}^{\mathrm{BP}}}
     {T_{\infty}^{\mathrm{TreeProp}}}
=
\frac{2N}{2D}
=
\frac{N}{D}.
\]

For the recurrent model experiments, we use $S_{\infty}
=
\frac{T_{\infty}^{\mathrm{BP}}}
     {T_{\infty}^{\mathrm{TreeProp}}}
=
\frac{2N}{1 + D + K}$ as the recurrent experiments use detached gradients making each gradient flow only one step, and we use $K = D$ as a conservative lower bound on the speed (all experiments use $K \le D$). This quantity represents an ideal algorithmic speedup and does not
account for communication, synchronization, memory-transfer, or
hardware-utilization overheads.
The experiments were conducted using NVIDIA GH200, H100, and H200 accelerators.
\subsection{MNIST Classification}
\label{app:mnist_details}
For MNIST, we use a simple multilayer perceptron with sigmoid activations and no residual connections or normalization layers.
We evaluate hierarchy depths $D \in \{2,3,4\}$, corresponding to networks with $4$, $8$, and $16$ layers, respectively.
For each depth, we vary the hidden dimensionality over
$\{50,100,500\}$.
The same architecture depth and hidden dimensionality are
used for the BP and CFF+M baselines whenever applicable.

For MNIST classification, we perform a grid search over batch sizes
$\{32,64\}$, learning rates
$\{10^{-2},10^{-3},10^{-4}\}$, and weight-decay values
$\{0.01,0.05\}$.

The exact hyperparameters for the MNIST classification are as follows: we use a batch size of 64, the AdamW optimizer \citep{loshchilov2018decoupled}, a learning rate of 0.001, weight-decay of 0.01 and train for 100 epochs.

For TreeProp, the sigmoid activations are interpreted as the mean parameters of Bernoulli latent variables. At each node, the propagated Bernoulli means are used to infer the corresponding local posterior distribution. 
The TreeProp posterior mean scales are fixed to $\alpha=\beta=0.5$ for all experiments.
These posterior means are propagated directly through the hierarchy, without sampling binary latent states or adding stochastic noise.

Table~\ref{tab:mnist_full_result} reports the full MNIST hidden-size sweep used in the main-text summary.
TreeProp remains competitive with BP for depths \(D=2\) and \(D=3\), while consistently outperforming CFF+M.
At depth \(D=4\), corresponding to 16-layer sigmoid MLPs, all BP
configurations collapse to chance-level performance of \(11.35\%\).
TreeProp remains trainable across all hidden sizes and achieves between
\(96.95\%\) and \(98.35\%\) accuracy, indicating its potential to cope with vanishing gradients, in line with later experiments.

\begin{table*}[t]
\centering
\small
\begin{tabular}{ccc|ccc}
\toprule
\multicolumn{3}{c|}{Model Configuration} & \multicolumn{3}{c}{Accuracy (\%)} \\
\cmidrule(lr){1-3} \cmidrule(lr){4-6}
Depth & Layers & Hidden Size & BP & CFF+M & TreeProp \\
\midrule
2 & 4  & 50  & $97.02 \pm 0.12$ & $95.07 \pm 0.08$ & $97.35 \pm 0.09$ \\
2 & 4  & 100 & $97.96 \pm 0.02$ & $95.52 \pm 0.02$ & $98.01 \pm 0.05$ \\
2 & 4  & 500 & $98.57 \pm 0.06$ & $96.26 \pm 0.05$ & $98.51 \pm 0.03$ \\
3 & 8  & 50  & $95.46 \pm 0.40$ & $95.08 \pm 0.11$ & $97.32 \pm 0.15$ \\
3 & 8  & 100 & $97.97 \pm 0.02$ & $95.60 \pm 0.02$ & $97.92 \pm 0.09$ \\
3 & 8  & 500 & $98.51 \pm 0.05$ & $96.12 \pm 0.05$ & $98.51 \pm 0.05$ \\
4 & 16 & 50  & $11.35 \pm 0.00$ & $95.00 \pm 0.07$ & $96.95 \pm 0.13$ \\
4 & 16 & 100 & $11.35 \pm 0.00$ & $95.60 \pm 0.03$ & $97.53 \pm 0.03$ \\
4 & 16 & 500 & $11.35 \pm 0.00$ & $95.98 \pm 0.03$ & $98.35 \pm 0.02$ \\
\bottomrule
\end{tabular}
\caption{MNIST classification results across hierarchy depths
\(D\in\{2,3,4\}\), corresponding to 4, 8, and 16-layer networks,
and hidden sizes \(\{50,100,500\}\). TreeProp remains trainable at
\(D=4\), where all evaluated 16-layer BP baselines collapse to
chance-level performance.}
\label{tab:mnist_full_result}
\end{table*}

\subsection{CIFAR-10 Classification}
\label{app:cifar10_details}
For CIFAR-10, we use a Vision Transformer architecture consisting of a
patch-embedding layer, Transformer blocks, and a classification head.
Each Transformer block uses an embedding dimension of \(240\) and six
attention heads.
The TreeProp model uses a depth-$3$ hierarchy.
The full path contains seven Transformer blocks, while the short path contains three Transformer blocks.

We perform a separate grid search for TreeProp, BP, and CFF+M and select
the best configuration for each method according to validation accuracy.
For TreeProp, we grid-search the learning rate over
\(\{10^{-3},5\times10^{-4},10^{-4}\}\), the batch size over
\(\{256,512\}\), the posterior variation scale over the evaluated values,
and the forward and right-boundary contribution coefficients over \(\alpha,\beta\in\{0,0.5,1\})\).
For BP, we separately grid-search the learning rate and batch size. For
CFF+M, we also perform a separate grid search, with the best-performing
configuration matching the hyperparameters provided in the original
work~\citep{constrative_fwd_fwd}.
TreeProp and BP are trained for \(500\) epochs. For CFF+M, we follow the
training schedule of the original work, training the
Transformer blocks for \(500\) epochs and the final classifier head for
\(100\) epochs. Each selected configuration is evaluated using seeds \(1\),
\(2\), and \(3\), and we report the mean and standard deviation of the
test accuracy over the three seeds.

The selected TreeProp configuration uses a batch size of \(512\), the
AdamW optimizer~\citep{loshchilov2018decoupled}, a learning rate of
\(5\times10^{-4}\), and a dropout rate of \(0.1\). Posterior samples are
obtained using the reparameterization trick with a posterior variation
scale of \(0.005\) and a depth-wise schedule, where the noise scale at
tree level \(d\) is multiplied by \(\sqrt{2^{D-d}}\), with \(D\)
denoting the hierarchy depth. The selected forward- and right-boundary
contribution coefficients are \(\alpha=1\) and \(\beta=0.\),
respectively.

The selected seven-layer BP baseline uses a batch size of \(512\), the
AdamW optimizer, a learning rate of \(10^{-3}\), and a dropout rate of
\(0.1\).

For CFF+M, we follow the two-stage training procedure of the reference
implementation. The Transformer blocks are first trained independently
using the contrastive objective. They are then frozen, and a classifier
head is trained on the resulting representations. All remaining CFF+M
settings follow the original work.


\subsection{CIFAR-100 Classification}
\label{app:cifar100_details}

For CIFAR-100, we use the same Vision Transformer architecture as in the
CIFAR-10 experiments, consisting of a patch-embedding layer, Transformer
blocks, and a classification head. Images are divided into \(4\times4\)
patches, and each Transformer block uses an embedding dimension of \(240\)
and six attention heads.

The TreeProp model uses a depth-\(3\) hierarchy. The full path contains
seven Transformer blocks, whereas the short path contains three
Transformer blocks. We compare against separately trained end-to-end
backpropagation baselines with the corresponding effective depths and
against our reproduction of CFF+M~\citep{constrative_fwd_fwd}.

We use the same data-augmentation pipeline for TreeProp, BP, and CFF+M,
consisting of random cropping, horizontal flipping, and
RandAugment~\citep{3495724.3497287}. We report top-\(1\) classification accuracy for all methods.

We perform a separate grid search for TreeProp, BP, and CFF+M and select
the best configuration for each method according to validation accuracy.
For TreeProp, we grid-search the learning rate, batch size, posterior
variation scale, and the forward- and right-boundary contribution
coefficients \(\alpha,\beta\in\{0,0.5,1\}\). For BP, we separately
grid-search the learning rate, batch size, weight decay, and dropout rate.
For CFF+M, we also perform a separate grid search, with the best-performing
configuration matching the hyperparameters provided in the original
work~\citep{constrative_fwd_fwd}. Each selected configuration is evaluated
using seeds \(1\), \(2\), and \(3\), and we report the mean and standard
deviation of the test accuracy over the three seeds.

The selected TreeProp configuration uses a batch size of \(512\), the
AdamW optimizer~\citep{loshchilov2018decoupled}, a learning rate of
\(5\times10^{-4}\), a weight decay of \(0.05\), a dropout rate of \(0.1\),
and \(500\) training epochs. Gradients are clipped to a maximum norm of
\(1.0\). Posterior samples are obtained using the reparameterization trick
with a posterior variation scale of \(0.002\) and a depth-wise schedule,
where the noise scale at tree level \(d\) is multiplied by
\(\sqrt{2^{D-d}}\), with \(D\) denoting the hierarchy depth.
The selected forward- and right-boundary
contribution coefficients are \(\alpha=1\) and \(\beta=0.\),
respectively.

The selected seven-layer BP configuration uses a batch size of \(512\),
the AdamW optimizer, a learning rate of \(5\times10^{-4}\), a weight decay
of \(0.05\), and no dropout, and \(500\) training epochs.
Gradients are clipped to a maximum norm of \(1.0\).

For CFF+M, we follow the two-stage training procedure of the reference
implementation. The Transformer blocks are first trained independently
using the contrastive objective. They are then frozen, and a classifier
head is trained on the resulting representations. The selected
configuration matches the hyperparameters provided in the original work,
and all remaining settings follow its reference implementation.

\subsection{TinyImageNet Classification}
\label{app:tinyimagenet_details}

For TinyImageNet, we use the same Vision Transformer architecture and
experimental protocol as in the CIFAR experiments. Each model uses an
embedding dimension of \(240\), six attention heads, and seven Transformer
blocks. Since TinyImageNet images have a resolution of \(64\times64\), we
divide them into \(8\times8\) patches. The TreeProp model uses a
depth-\(3\) hierarchy, where the full path contains seven Transformer
blocks and the short path contains three Transformer blocks.

We compare TreeProp against separately trained end-to-end backpropagation
baselines with the corresponding effective depths and against our
reproduction of CFF+M~\citep{constrative_fwd_fwd}. We use the same
data-augmentation pipeline for all methods, consisting of random cropping,
horizontal flipping, and RandAugment~\citep{3495724.3497287}. The
TinyImageNet contains \(200\) classes, and we report top-\(1\) classification
accuracy.

As in the CIFAR experiments, we perform a separate grid search for
TreeProp, BP, and CFF+M and select the best configuration for each method
according to validation accuracy. Each selected configuration is evaluated
using seeds \(1\), \(2\), and \(3\), and we report the mean and standard
deviation of the test accuracy over the three seeds.

The selected TreeProp configuration uses a batch size of \(512\), the
AdamW optimizer~\citep{loshchilov2018decoupled}, a learning rate of
\(5\times10^{-4}\), a weight decay of \(0.05\), a dropout rate of \(0.1\),
and \(500\) training epochs. Gradients are clipped to a maximum norm of
\(1.0\). Posterior samples are obtained using the reparameterization trick
with a posterior variation scale of \(0.005\) and a depth-wise schedule,
where the noise scale at tree level \(d\) is multiplied by
\(\sqrt{2^{D-d}}\), with \(D\) denoting the hierarchy depth.
The selected forward- and right-boundary
contribution coefficients are \(\alpha=1\) and \(\beta=0.\),
respectively.

The selected seven-layer BP configuration uses a batch size of \(512\),
the AdamW optimizer, a learning rate of \(5\times10^{-4}\), a weight decay
of \(0.05\), a dropout rate of \(0.1\), and \(500\) training epochs.
Gradients are clipped to a maximum norm of \(1.0\).

For CFF+M, we use the same two-stage training procedure and selected
hyperparameter configuration as in the CIFAR experiments. The Transformer
blocks are first trained independently using the contrastive objective.
They are subsequently frozen, and a classifier head is trained on the
resulting representations. The selected configuration uses a batch size
of \(512\), stage-wise learning rates of \(4\times10^{-3}\) and
\(5\times10^{-4}\), a contrastive-loss temperature of \(0.15\), and a
linear margin schedule from \(m_0=0.4\) to \(m_L=0.1\). All remaining
settings follow the reference implementation.

\subsection{Language Modeling on WikiText-103}
\label{app:language_modeling_details}

\begin{table}[t]
\centering
\small
\setlength{\tabcolsep}{4pt}
\begin{tabular}{@{}lccc@{}}
\toprule
& \multicolumn{3}{c}{Depth \(D\) / Transformer blocks \(N\)} \\
\cmidrule(lr){2-4}
& \(4/12\) & \(5/24\) & \(6/36\) \\
\midrule

Hidden size \(d\)
& 768 & 1024 & 1280 \\

Attention heads
& 12 & 16 & 20 \\

Batch per GPU
& 8 & 8 & 6 \\

Global batch
& 32 & 32 & 24 \\

Steps per epoch
& 3595 & 3595 & 4794 \\

Peak learning rate
& \(4\times10^{-4}\)
& \(4\times10^{-4}\)
& \(3\times10^{-4}\) \\

LR schedule
& linear
& linear
& constant \\

Warmup ratio
& \(0.00174\)
& \(0.00174\)
& \(0.01\) \\

Epoch budget
& 30 & 40 & 20 \\

\bottomrule
\end{tabular}
\caption{Hyperparameters used for the backpropagation models and the
corresponding TreeProp configurations on WikiText-103.}
\label{tab:lm-hparams}
\end{table}

\begin{table}[t]
\centering
\small
\setlength{\tabcolsep}{4pt}

\resizebox{\columnwidth}{!}{%
\begin{tabular}{lc|ccc}
\toprule
Dataset & $\alpha \backslash \beta$
& $0$ & $0.50$ & $1.00$ \\
\midrule

\multirow{3}{*}{CIFAR-10}
& $0$
& --
& $13.38 \pm 0.18$
& $13.23 \pm 0.77$ \\
& $0.50$
& $87.92 \pm 0.44$
& $88.46 \pm 0.21$
& $88.32 \pm 0.21$ \\
& $1.00$
& $88.71 \pm 0.35$
& $88.54 \pm 0.25$
& $88.18 \pm 0.23$ \\
\midrule

\multirow{3}{*}{CIFAR-100}
& $0$
& --
& $1.16 \pm 0.06$
& $1.07 \pm 0.04$ \\
& $0.50$
& $62.03 \pm 0.34$
& $62.28 \pm 0.64$
& $62.11 \pm 0.37$ \\
& $1.00$
& $62.69 \pm 0.49$
& $62.35 \pm 0.58$
& $62.45 \pm 0.81$ \\
\midrule

\multirow{3}{*}{TinyImageNet}
& $0$    & -- & $0.53 \pm 0.17$ & $0.50 \pm 0.05$ \\
& $0.50$ & $44.61 \pm 0.28$  & $44.23 \pm 0.21$ & $44.31 \pm 0.44$ \\
& $1.00$ & $45.23 \pm 0.32$ & $44.79 \pm 0.20$ & $44.78 \pm 0.41$ \\
\bottomrule
\end{tabular}%
}

\caption{Top-\(1\) classification accuracy (\%) for different values of
the forward and feedback scales \(\alpha\) and \(\beta\). Results report
the mean and sample standard deviation over three seeds. The configuration
\(\alpha=\beta=0\) is excluded.}
\label{tab:alpha_beta_ablation}
\end{table}

\begin{table*}[ht]
\centering
\small
\setlength{\tabcolsep}{4pt}
\begin{tabular}{@{}llc|ccc@{}}
\toprule
Dataset
& TreeProp subnetwork
& Transformer blocks
& Final CE only
& Node losses \(+\) final CE
& Node losses \(+\) all CE \\
\midrule

\multirow{4}{*}{CIFAR-10}
& \(\bx\!\rightarrow\!\bz_4\!\rightarrow\!\by\)
& 1
& \(20.21 \pm 3.41\)
& \(16.06 \pm 1.51\)
& \(78.60 \pm 0.21\) \\

& \(\bx\!\rightarrow\!\bz_4\!\rightarrow\!\bz_6
  \!\rightarrow\!\by\)
& 2
& \(39.39 \pm 2.68\)
& \(23.37 \pm 4.04\)
& \(85.91 \pm 0.06\) \\

& \(\bx\!\rightarrow\!\bz_4\!\rightarrow\!\bz_6
  \!\rightarrow\!\bz_7\!\rightarrow\!\by\)
& 3
& \(56.78 \pm 11.59\)
& \(89.08 \pm 0.09\)
& \(87.36 \pm 0.11\) \\

& \(\bx\!\rightarrow\!\bz_1\!\rightarrow\!\cdots
  \!\rightarrow\!\by\)
& 7
& \(28.16 \pm 0.09\)
& \(88.54 \pm 0.25\)
& \(86.83 \pm 0.08\) \\

\midrule

\multirow{4}{*}{CIFAR-100}
& \(\bx\!\rightarrow\!\bz_4\!\rightarrow\!\by\)
& 1
& \(9.61 \pm 5.71\)
& \(2.44 \pm 1.04\)
& \(50.54 \pm 0.71\) \\

& \(\bx\!\rightarrow\!\bz_4\!\rightarrow\!\bz_6
  \!\rightarrow\!\by\)
& 2
& \(21.51 \pm 6.08\)
& \(8.76 \pm 2.62\)
& \(58.18 \pm 0.40\) \\

& \(\bx\!\rightarrow\!\bz_4\!\rightarrow\!\bz_6
  \!\rightarrow\!\bz_7\!\rightarrow\!\by\)
& 3
& \(40.60 \pm 18.67\)
& \(63.22 \pm 0.55\)
& \(59.04 \pm 0.40\) \\

& \(\bx\!\rightarrow\!\bz_1\!\rightarrow\!\cdots
  \!\rightarrow\!\by\)
& 7
& \(26.15 \pm 30.39\)
& \(62.35 \pm 0.58\)
& \(58.17 \pm 0.40\) \\

\midrule

\multirow{4}{*}{TinyImageNet}
& \(\bx\!\rightarrow\!\bz_4\!\rightarrow\!\by\)
& 1
& \(4.17 \pm 1.47\)
& \(2.93 \pm 0.19\)
& \(35.36 \pm 0.30\) \\

& \(\bx\!\rightarrow\!\bz_4\!\rightarrow\!\bz_6
  \!\rightarrow\!\by\)
& 2
& \(14.35 \pm 2.33\)
& \(11.79 \pm 1.02\)
& \(41.30 \pm 0.21\) \\

& \(\bx\!\rightarrow\!\bz_4\!\rightarrow\!\bz_6
  \!\rightarrow\!\bz_7\!\rightarrow\!\by\)
& 3
& \(39.10 \pm 10.64\)
& \(45.69 \pm 0.12\)
& \(41.80 \pm 0.15\) \\

& \(\bx\!\rightarrow\!\bz_1\!\rightarrow\!\cdots
  \!\rightarrow\!\by\)
& 7
& \(31.57 \pm 22.64\)
& \(44.79 \pm 0.20\)
& \(41.41 \pm 0.21\) \\

\bottomrule
\end{tabular}

\caption{Ablation of the TreeProp training objective across tree-induced
subnetworks. We compare training with only the final cross-entropy loss,
with node-level losses and the final cross-entropy loss, and with the
complete objective containing node-level losses and cross-entropy losses
at all the tree levels. Results report the mean and sample standard
deviation of top-\(1\) test accuracy (\%) over three seeds.}
\label{tab:objective_path_ablation}
\end{table*}

For language modeling, we use a GPT-2-style Transformer architecture based on \citet{karpathy2022mingpt}.
Each TreeProp block corresponds to one Transformer
block. 
We evaluate architectures matching GPT-2 Small, Medium, and Large.
GPT-2 Small uses \(12\) Transformer blocks, \(12\) attention heads, and
an embedding dimension of \(768\); GPT-2 Medium uses \(24\) blocks,
\(16\) attention heads, and an embedding dimension of \(1024\); and
GPT-2 Large uses \(36\) blocks, \(20\) attention heads, and an embedding
dimension of \(1280\). The corresponding TreeProp models match these
architectural configurations, including the number of Transformer blocks,
attention heads, and embedding dimension.
Performance is measured using perplexity on WikiText-103. 
For all the language-modeling experiments, WikiText-103 is tokenized using the pretrained GPT-2 byte-level BPE tokenizer provided by the Hugging Face Transformers library. 
The model-specific architecture and optimization
hyperparameters are summarized in Table~\ref{tab:lm-hparams}.
In all the experiments we use mixed-precision training with bfloat16.
The tokenized sequences are concatenated and divided into non-overlapping
context windows of \(1024\) tokens.
Due to computational-resource constraints, we perform a limited
hyperparameter sweep and select configurations based on validation
perplexity after the first training epoch. The sweep includes the learning
rate and the remaining selected training hyperparameters.

For the TreeProp language model, posterior samples are obtained using the
reparameterization trick with a posterior variation scale of \(0.005\)
and a depth-wise schedule. At tree level \(d\), the noise scale is
multiplied by \(\sqrt{2^{D-d}}\), where \(D\) denotes the hierarchy depth. We use \(\alpha=1\) and \(\beta=0.5\) to weight the forward estimate and the right-boundary correction, respectively.

We do not report CFF+M in this section because we could not identify a directly comparable autoregressive language-modeling result for this setting.

\subsection{(Permuted-)Sequential MNIST Classification (TreeProp-TT)}
\label{app:smnist_details}
For sequential and permuted-sequential MNIST we use the recurrent variant of TreeProp (TreeProp-TT),
whose $N$ time steps are arranged in a binary tree and trained block-locally; the deployed model is
the single depth-$0$ cell rolled over the full sequence as an ordinary RNN. We evaluate two cells: a
gated GRU \citep{cho-etal-2014-learning} and a gateless Elman cell (a single linear layer with
sigmoid activation) \citep{ELMAN1990179}, both at a fixed
hidden width of $2048$. We sweep the tree depth $D$, with sequence length $N{=}2^{D}$; we report
$D \in \{1,\dots,8\}$ for the GRU and $D \in \{1,\dots,6\}$ for the gateless Elman cell. Each image is flattened to its $784$ pixels and, at depth
$D$, split into $N$ steps of $\lceil 784/2^{D}\rceil$ consecutive pixels; the pixels are start-padded
with leading zeros to $N\lceil 784/2^{D}\rceil$, the minimal padding that makes the split even and
leaves the deployed chain ending on real pixels (no padding for $D\leq 4$, since $784=2^{4}\cdot 49$).
Permuted-sequential MNIST additionally applies a single fixed random permutation to the $784$ pixels
before padding, shared across train and test. The same cell type, hidden width, and parameter count are used for the
BPTT baseline, which is trained by full $\mathcal{O}(N)$ backpropagation through time.

The exact hyperparameters for the (permuted-)sequential MNIST experiments are as follows: we use a
batch size of $128$, the AdamW optimizer \citep{loshchilov2018decoupled}, a learning rate of $0.001$,
weight decay of $10^{-4}$, gradient-norm clipping at $5.0$, and train for $200$ epochs. We use
mixed-precision training with bfloat16. TreeProp-TT is trained block-locally at $\mathcal{O}(\log N)$
span, with all nodes at a given depth evaluated in one batched call and inter-node states detached.
The deployed chain is additionally trained with the parallel fixed-point (consistency) correction at
consistency weight $1.0$, differentiating only through the final sweep; we select the number of
iterations $K$ per depth (the best $K$ reported in each table, with $K{=}0$ denoting tree-only, no
correction, which is used throughout for the gateless Elman cell). Activations are treated
deterministically without added noise, and a single recurrent cell is shared per depth in the tree so that
the deployed depth-$0$ chain is one cell rolled. For the timing comparisons we implement the GRU cell explicitly, as plain matrix multiplications, for both TreeProp and the BPTT baseline, rather than using PyTorch's \texttt{nn.GRU}. The latter dispatches to a fused cuDNN kernel, which would give the baseline a kernel-level speed advantage unrelated to the training algorithm; since TreeProp does not yet have an equivalent fused CUDA kernel, using the unfused cell for both sides isolates the algorithmic (training-span) speedup and keeps the comparison fair.


\subsection{Ablation studies}
\label{app:ablation}

\subsubsection{Ablation of posterior-mean weighting.}

\begin{table*}[t]
\centering
\small
\setlength{\tabcolsep}{3.5pt}
\begin{tabular}{l|cccccccccc}
\toprule
& \multicolumn{10}{c}{Tree depth \(D\)} \\
\cmidrule(lr){2-11}
Setting & 1 & 2 & 3 & 4 & 5 & 6 & 7 & 8 & 9 & 10 \\
\midrule
\multicolumn{11}{@{}l}{\emph{sequential MNIST}} \\
\(K{=}0\) & 98.6 & 98.5 & 98.4 & 98.5 & 98.4 & 98.2 & 97.8 & 97.5 & 46.1 & 18.9 \\
\(K{=}4\) & 98.7 & 98.6 & 98.6 & 98.7 & 98.6 & 98.3 & 98.1 & 97.7 & 97.2 & 94.8 \\
\addlinespace
\multicolumn{11}{@{}l}{\emph{permuted-sequential MNIST}} \\
\(K{=}0\) & 98.6 & 98.5 & 98.4 & 98.2 & 98.1 & 97.3 & 97.3 & 95.0 & 56.5 & 70.5 \\
\(K{=}4\) & 98.7 & 98.5 & 98.6 & 98.5 & 98.3 & 97.9 & 96.9 & 95.1 & 90.1 & 83.4 \\
\bottomrule
\end{tabular}
\caption{Effect of the consistency correction on the deployed depth-\(0\)
chain. Test accuracy (\%) of the rolled-out cell, GRU, \(H{=}2048\), 200
epochs. \(K{=}0\) trains the tree only; \(K{=}4\) adds the parallel
fixed-point correction. The correction is immaterial for \(D\leq 8\) and
decisive at \(D\geq 9\).}
\label{tab:rnn_k_ablation}
\end{table*}

We evaluate the effect of the coefficients $\alpha$ and $\beta$, which control the contributions of the forward estimate and the right-boundary correction used to calculate the posterior mean in Eq.~\eqref{eqn:posterior-mean} respectively. The same coefficient configurations are tested on CIFAR-10, CIFAR-100, and TinyImageNet, while retaining the dataset-specific architecture and training settings used in the main experiments. Table~\ref{tab:alpha_beta_ablation} summarizes the resulting classification accuracies.

\subsubsection{Effect of Node-wise losses and intermediate CE losses}
\label{app:abl-nodewise-losses}
We evaluate how the additional KL and intermediate cross-entropy losses affect the subnetworks induced by different paths through the TreeProp hierarchy. For each dataset, we report the different subnetworks under three objective configurations: final cross-entropy only, all node-level losses with final cross-entropy in Eq ~\eqref{eqn:tree-variational-objective}, and the objective  with all node-wise losses and cross-entropy losses at all eligible tree levels. All TreeProp subnetworks are extracted from the corresponding jointly trained model, while each BP result is obtained from a separately trained end-to-end model with the same effective number of Transformer blocks.
Table ~\ref{tab:objective_path_ablation} shows that training with all
node-wise losses, the final cross-entropy loss, and intermediate
cross-entropy losses at all eligible tree levels enables TreeProp to
learn subnetworks with comparable performance across different
inference paths. 
When the primary objective is to maximize the
performance of the longest path, the intermediate cross-entropy losses
can instead be omitted during training.


We conduct a series of ablation studies to examine the main design
choices in TreeProp and to separate their contributions to optimization
and subnetwork performance.
In particular, we study the effects of the
node-wise variational losses, the intermediate cross-entropy terms, and
the relative weighting of the forward estimate and right-boundary
correction through the coefficients \(\alpha\) and \(\beta\).
We also evaluate the subnetworks induced by different paths through the jointly
trained hierarchy.
Unless stated otherwise, all comparisons use the same
architecture and training protocol explained in ~\ref{app:experimental-details}, and results are reported as the mean
and standard deviation over three random seeds.

\subsubsection{Ablations for the Recurrent Variant (TreeProp-TT)}
\label{app:rnn_ablation}

We ablate the two free choices of the recurrent setup described in
Appendix~\ref{app:smnist_details}: the number of fixed-point (consistency)
sweeps \(K\) and the weight \(\lambda_{\mathrm{cons}}\) of the consistency
objective in Eq.~\eqref{eq:consistency_loss}. All runs use the GRU cell at
hidden width \(H{=}2048\) and the protocol of
Appendix~\ref{app:smnist_details}; depths \(D{=}9,10\) extend the range
reported in Table~\ref{tab:TreeProp-all} and are included because the effect of
the correction is concentrated there. Each cell is a single run, so
differences below roughly \(0.5\) points should be read as no measurable
effect.

\paragraph{Number of consistency sweeps \(K\).}
Table~\ref{tab:rnn_k_ablation} contrasts the uncorrected tree
(\(K{=}0\)) with the corrected chain (\(K{=}4\)). Up to \(D{=}8\) the
correction changes the deployed accuracy by at most a few tenths of a point in
either direction, which is why several entries in Table~\ref{tab:TreeProp-all}
select \(K{=}0\). Beyond that the picture changes completely: at \(D{=}9\) and
\(D{=}10\) the uncorrected chain collapses, and the correction recovers
\(51.1\) and \(75.9\) points respectively on sequential MNIST. This is the
exposure-bias mechanism of Appendix~\ref{app:recurrent_TreeProp} becoming
dominant once the rollout is long enough, and it is the reason \(K\) is
selected per depth rather than fixed globally.

Increasing \(K\) further yields no consistent gain while its cost grows
linearly: at \(D{=}9\) accuracy improves by \(2.9\) points from \(K{=}4\) to
\(K{=}6\) and then flattens, and at \(D{=}10\) it varies non-monotonically with
no net improvement, while epoch time rises from \(90.7\)\,s at \(K{=}4\) to
\(127.9\)\,s at \(K{=}8\).

\begin{table}[t]
\centering
\small
\setlength{\tabcolsep}{4pt}
\begin{tabular}{l|cccc}
\toprule
& \(K{=}0\) & \(K{=}4\) & \(K{=}6\) & \(K{=}8\) \\
\midrule
Acc., \(D{=}9\)   & 56.5 & 90.1 & 93.0 & 92.8 \\
Acc., \(D{=}10\)  & 70.5 & 83.4 & 76.6 & 83.4 \\
\bottomrule
\end{tabular}
\caption{Sweep over the number of consistency iterations \(K\) at the depths
where the correction matters (permuted-sequential MNIST, GRU).}
\label{tab:rnn_k_sweep}
\end{table}

\paragraph{Consistency weight.}
Table~\ref{tab:rnn_weight_ablation} varies \(\lambda_{\mathrm{cons}}\) with all
else fixed. The reported value \(\lambda_{\mathrm{cons}}{=}1.0\) is best at both
depths tested, and the gap widens at \(D{=}9\), where down-weighting the
correction costs \(3.5\) points and up-weighting it costs \(0.9\).

\begin{table}
\centering
\small
\begin{tabular}{l|ccc}
\toprule
& \(\lambda_{\mathrm{cons}}{=}0.5\) & \(\lambda_{\mathrm{cons}}{=}1.0\) & \(\lambda_{\mathrm{cons}}{=}2.0\) \\
\midrule
\(D{=}8\) & 97.6 & \textbf{97.7} & 97.3 \\
\(D{=}9\) & 93.7 & \textbf{97.2} & 96.2 \\
\bottomrule
\end{tabular}
\caption{Effect of the consistency weight on the deployed chain (sequential
MNIST, GRU, \(K{=}4\)). Test accuracy (\%).}
\label{tab:rnn_weight_ablation}
\end{table}

\bibliography{refs}
\end{document}

%% file: math_symbols.tex
\usepackage{amsmath}        
\usepackage{amsthm}

\newcommand{\ba}{\mathbf{a}}

\newcommand{\bx}{\mathbf{x}}
\newcommand{\by}{\mathbf{y}}
\newcommand{\bz}{\mathbf{z}}
\newcommand{\beps}{\boldsymbol{\varepsilon}}

\newtheorem{theorem}{Theorem}
\newtheorem{theorem-supp}{Theorem}[section]